\documentclass[10pt,twocolumn,letterpaper]{article}

\usepackage{cvpr}              % To produce the CAMERA-READY version
\usepackage[dvipsnames]{xcolor}

\definecolor{cvprblue}{rgb}{0.21,0.49,0.74}
\usepackage[pagebackref,breaklinks,colorlinks,citecolor=cvprblue]{hyperref}

\usepackage{multirow}
\usepackage{subfigure}

\usepackage{algorithm}
\usepackage{algorithmic}
\usepackage[accsupp]{axessibility}

\usepackage{amsmath}
\usepackage{amssymb}
\usepackage{mathtools}
\usepackage{amsthm}
\usepackage{soul}
\usepackage[capitalize,noabbrev]{cleveref}

\theoremstyle{plain}
\theoremstyle{definition}
\newtheorem{definition}{Definition}

\theoremstyle{remark}

\newtheoremstyle{TheoremRep}
        {\topsep}{\topsep}              %%% space between body and thm
        {\itshape}                      %%% Thm body font
        {}                              %%% Indent amount (empty = no indent)
        {\bfseries}                     %%% Thm head font
        {.}                             %%% Punctuation after thm head
        { }                             %%% Space after thm head
        {\thmname{#1}\thmnote{ \bfseries #3}}%%% Thm head spec
\theoremstyle{TheoremRep}

\newcommand{\E}{\ensuremath{\mathbb{E}}}

\newcommand{\Y}{\ensuremath{\mathcal{Y}}}
\newcommand{\X}{\ensuremath{\mathcal{X}}}
\newcommand{\Z}{\ensuremath{\mathcal{Z}}}

\def\paperID{18} % *** Enter the Paper ID here
\def\confName{CVPR}
\def\confYear{2024}

\title{Test-time Assessment of a Model's Performance on Unseen Domains \\ via Optimal Transport}

\author{Akshay Mehra\textsuperscript{1}, Yunbei Zhang\textsuperscript{1}, and Jihun Hamm\textsuperscript{1}\\
{\small \textsuperscript{1}Tulane University}\\ 
{\tt\small\{amehra, yzhang111, jhamm3\}@tulane.edu}\\
}
\begin{document}
\maketitle
\begin{abstract}
Gauging the performance of ML models on data from unseen domains at test-time is essential yet a challenging problem due to the lack of labels in this setting. 
Moreover, the performance of these models on in-distribution data is a poor indicator of their performance on data from unseen domains.
Thus, it is essential to develop metrics that can provide insights into the model's performance at test time and can be computed only with the information available at test time (such as their model parameters, the training data or its statistics, and the unlabeled test data).
To this end, we propose a metric based on Optimal Transport that is highly correlated with the model's performance on unseen domains and is efficiently computable only using information available at test time. 
Concretely, our metric characterizes the model's performance on unseen domains using only a small amount of unlabeled data from these domains and data or statistics from the training (source) domain(s).
Through extensive empirical evaluation using standard benchmark datasets, and their corruptions, we demonstrate the utility of our metric in estimating the model's performance in various practical applications. 
These include the problems of selecting the source data and architecture that leads to the best performance on data from an unseen domain and the problem of predicting a deployed model's performance at test time on unseen domains.
Our empirical results show that our metric, which uses information from both the source and the unseen domain, is highly correlated with the model's performance, achieving a significantly better correlation than that obtained via the popular prediction entropy-based metric, which is computed solely using the data from the unseen domain.

%Machine learning (ML) models deployed in the real world often encounter data from distributions different from the ones they have been trained on.
%In such scenarios, the performance of these models on in-distribution data does not correlate well with their performance on out-of-distribution data. 
%Moreover, ML models can access limited information (such as their parameters, the training data or its statistics, and the unlabeled test data) at test time for making predictions.
%Thus it is essential to develop metrics that can be computed without the target domain's labels but are still effective for gauging model performance in this setting.
\end{abstract}

\section{Introduction}
\begin{figure*}
\small
\centering
\includegraphics[width=0.9\linewidth]{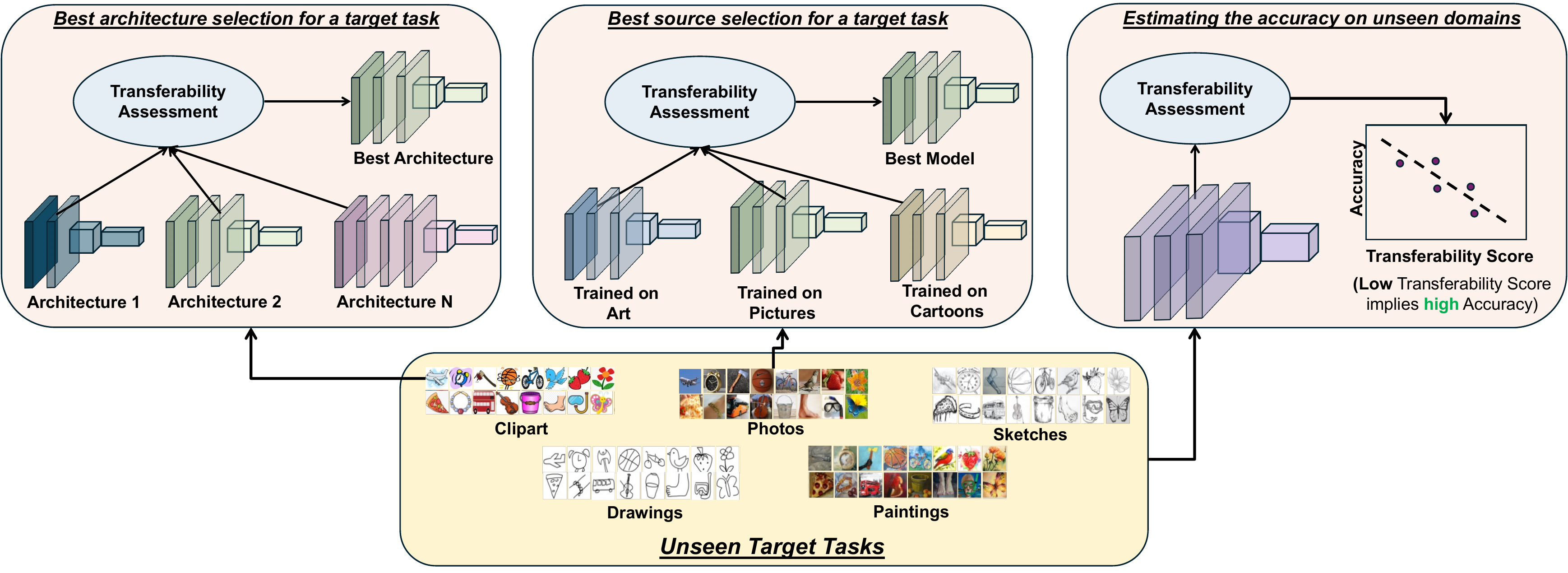}
\caption{(Best viewed in color.) Overview of the practical applications for which TETOT can be utilized. 
The first application (left) is to identify the model architecture that will yield the highest transferability for a particular target domain.
The second application (center) is to identify the best source domain data that will produce a model with the highest transferability to the target domain.
The third application (right) is to assess the performance of a given model on unseen target domains given only unlabeled data from those domains.
}
\label{fig:transferability_applications}
\end{figure*}

Machine Learning (ML) models deployed in the real world are often faced with data from distributions that differ significantly from the ones used for model training.
In such scenarios, prior works \cite{hendrycks2019robustness,bulusu2020anomalous} have shown a significant degradation in the performance of these models.
This makes the model's performance on data from training distribution a poor indicator of their performance on real-world distributions.
Thus, in this work, we focus on the problem of developing a metric that can be used at test time and can assess the model's performance on unseen domains in a cross-domain setting (referred to as transferability, Def~\ref{def:transferability}, in our work). 
This setting is similar to that used in the domain adaptation (DA) \cite{ben2010theory,mansour2009domain} and domain generalization (DG) \cite{wang2021generalizing} literature where the tasks are the same across domains (e.g., digit classification for MNIST and SVHN).
Such a metric is essential for various practical applications such as finding the best pre-trained model (trained with different sources and model architectures) at test time, that yields the best performance on an unseen target domain. 
It can also gauge the performance of a deployed model on unseen target domains at test time without requiring labels from these domains. 
See Fig.~\ref{fig:transferability_applications} for an overview of these applications.
%We give an overview of these applications in Fig.~\ref{fig:transferability_applications}. 

Since this metric will be used at test time to estimate the model's performance, it must be efficiently computable only using information available at test time, which includes access to unlabeled data from the target domain, parameters of the pre-trained models, and the knowledge of the source data (or its statistics) used for training the models. % and should be highly predictive of transferability.
Thus, we propose, TETOT (Test-time Estimation of Transferability via Optimal Transport) a metric that quantifies transferability in terms of the distributional divergence between the distribution of the source and that of the unseen target domain.
Specifically, we use Optimal Transport (OT) to estimate the distributional divergence between the distributions of the two domains and show that it is highly predictive of the transferability of the models. %in previously mentioned applications.
A series of analytical works \cite{ben2007analysis,albuquerque2019generalizing,mehra2021understanding,zhao2019learning,shen2018wasserstein,courty2017joint} in both DA and DG has highlighted the role of distributional distance-based measures such as Wasserstein distance, KL divergence, total variation distance, etc., in estimating model performance under distribution shifts.
While previous works in DA and DG \cite{ganin2016domain,long2018conditional,albuquerque2019generalizing,gulrajani2020search} focus on training models that are robust to distribution shifts, by minimizing distributional divergence, we focus on proposing a metric to estimate the transferability of pre-trained models at test time, useful for various practical applications (Fig.~\ref{fig:transferability_applications}).
% previous works have not focused on developing metrics to assess the transferability of models 

TETOT is efficiently computable at test time using unlabeled samples from the unseen domain. 
In particular, TETOT uses a small number of labeled samples from the source (training) distribution and uses the source classifier to obtain the pseudo-labels for the target domain samples. 
%Using these, TETOT can be efficiently estimated. % and is highly correlated with transferability. 
Moreover, for scenarios, where the source domain data is not accessible (e.g., due to privacy constraints) we also propose a variant of TETOT that can be estimated by using only statistics (mean and covariance) of the source data. 
The efficient computability of TETOT with information accessible at test time makes it practically useful for gauging the model's performance in this setting. 

To evaluate the effectiveness of TETOT we evaluate the Pearson correlation coefficient between transferability and TETOT on various practical problems. 
We use the popular PACS and VLCS benchmark datasets along with their corrupted versions in our evaluation and consider models trained using both a single domain and multiple domains. 
We evaluate TETOT on the problems of architecture and source domain selection. 
These problems aim to find the model architecture and training data that will yield the highest performance transferability to the data from an unseen domain.
We also evaluate the effectiveness of TETOT on the problem of estimating the transferability of a model on various unseen domains. 
Our empirical results demonstrate that TETOT produces a high correlation with transferability in all the applications, making it an effective metric for estimating transferability at test time.

We extensively compare the correlation of TETOT with transferability with average prediction entropy, a popular metric that is also applicable in the test time setting considered here but does not require the information of the source data for its computation. 
This metric has been shown to achieve a good correlation with transferability in \cite{wang2020tent} and has been used extensively to adapt the model to distribution shifts by using test-time adaptation \cite{wang2020tent,mummadi2021test,thopalli2023domain,liang2020we,iwasawa2021test}.
Even though prediction entropy is highly effective in predicting transferability, TETOT significantly outperforms it in all applications. 
This suggests the importance of utilizing source domain information in metrics to compute transferability. 
Our main contributions are summarized below:

\begin{itemize}[leftmargin=0.6cm]
    \item We propose TETOT, for assessing the transferability of classification models to unseen domains at test time based on the divergence between the distributions of the source and the target domains.
    
    \item TETOT is efficiently computable using unlabeled target domain data, requires only a small number of samples from the source, and can even be computed with access to only statistics of the source data.
    
    \item We demonstrate the effectiveness of TETOT in producing a better correlation with transferability compared to the popular entropy-based metric on practical problems including architecture selection, source selection, and estimating model performance on unseen domains. %We also demonstrate TETOT's superiority over prediction entropy in achieving a better correlation with transferability.
    
\end{itemize}

% This problem is extensively studied in the area of domain adaptation \cite{ben2010theory,mansour2009domain}
% and domain generalization literature intending to develop models resilient to distribution shifts. 
% Analytical works \cite{ben2007analysis,albuquerque2019generalizing,mehra2021understanding,zhao2019learning,shen2018wasserstein,courty2017joint} in these areas have emphasized the importance of distributional divergence between the training and the unseen domain's distribution as a crucial factor for explaining model's performance on unseen domains.

% Most UDA methods aim to minimize the distance-based metric but not many have shown it's correlation towards predicting transferability on the 3 applications. Moreover, re-training or fine-tuning models is becoming hard because of their size making it important to develop methods for test time analysis and adaptation. 

\section{Related Work}
\label{sec:related_work}

{\bf Learning under distribution shifts:}
ML models suffer degradation of performance when faced with data from unseen distributions. 
Analytical works in the areas of DA \cite{ben2007analysis,ben2010theory,mansour2009domain,shen2018wasserstein,mehra2021understanding,zhao2019learning} and DG \cite{albuquerque2019generalizing,zhang2021quantifying,ganin2016domain,zhao2018adversarial,qiao2020learning,gulrajani2020search} explain the reason for performance degradation in terms of the divergence between the distributions used for training and that encountered at test time. 
Following these, a large body of works \cite{long2018conditional,hoffman2018cycada,ganin2016domain,zhao2019learning} exists that learn models by minimizing different divergence metrics for learning a representation space where the distance between the distributions of training and unseen distributions can be minimized. 
Based on these, we propose to use a distance-based metric dependent on both source and unseen domain data to get insights into the performance of a model at test time allowing us to select the best model from a set of models (having different architectures and trained on different datasets) to use for prediction.
\\
\noindent{\bf Prediction entropy for transferability estimation:}
Recently, \cite{wang2020tent,mummadi2021test,thopalli2023domain,liang2020we,iwasawa2021test} showed the effectiveness of prediction entropy as a metric to adapt models at test time to distribution shift.
Two main factors make prediction entropy suitable for this task. First, it is correlated well with transferability to unseen domains (especially common corruptions of benchmark datasets \cite{hendrycks2018benchmarking,mehra2024fly}) and second, the ease of its estimation only using unlabeled target domain data. 
However, as we demonstrate extensively in Sec.~\ref{sec:experiments}, our TETOT metric is much better correlated with transferability and only has a slightly higher computational cost compared to entropy in our experiments. 
%can be computed with a similar computational cost as the entropy. 
\\
\noindent{\bf Estimating performance after transfer learning (TL):}
A recent line of work in TL focuses on developing scores \cite{8803726,nguyen2020leep,huang2022frustratingly,you2021logme,tan2021otce,mehra2023analysis} correlated with the performance of models after they have been fine-tuned on the target domain (referred to as full fine-tuning in TL). 
Some of the works in this area include Negative Conditional Entropy (NCE) \cite{tran2019transferability}, which uses negative conditional entropy between the true labels of the source and target domains, LEEP \cite{nguyen2020leep}, which computes NCE using the target domain labels and pseudo labels for the target domain from a source pre-trained model, and  OTCE \citep{tan2021otce}, which uses optimal transport coupling to compute conditional entropy between the two domains, to estimate performance after full fine-tuning. 
While these works show the utility of their metrics on the problems of source model/architecture selection, our work and setting are fundamentally different from works in this line of research. 
This is due to the following reasons. 
Firstly, we work in the setting when the label sets of the source and target domains are the same (see Sec.~\ref{sec:problem_setting}) unlike in TL where label sets could be different.
Secondly, we assume that we do not have access to the labels of the target domain (which is a critical requirement for estimating the scores proposed in this line of work).
Lastly, we define transferability as the accuracy of the pre-trained model on the target domain (Def.~\ref{def:transferability}) rather than the performance after fine-tuning using labels of the target domains, as used in this line of work.

\section{Test-time assessment of transferability}
Here we present our definition of the transferability of a model trained on source domain(s) to an unseen target domain. 
This is followed by the details of our proposed metric, TETOT, for estimating transferability at test time. 

\subsection{Notation and problem setting}
\label{sec:problem_setting}
Let $P_S(x, y)$ and $P_T(x, y)$ denote the distributions of the source and the target domains defined on $\X_S \times \Y_S$ and $\X_T \times \Y_T$ respectively. 
Let $\mathcal{D}_S^i = {(x_j^i, y_j^i)}_{j=1}^{m^i} \sim P_S^i(x,y)$  denote the $m^i$ samples from the $i^{th}$ source domain with $i \in \{1, ..., N_S\}$ where $N_S$ denotes the number of source domains used for training the model. 
Similarly, let $\mathcal{D}_T = {(x_j, y_j)}_{j=1}^{n} \sim P_T(x,y)$ denote $n$ samples from the target domain.
We assume that the feature spaces are common (i.e., $\X_S=\X_T=\X$) such as RGB images of the same input dimension but from different domains (such that $P_S \neq P_T$), and the same target label sets (i.e., $\Y_S = \Y_T = \Y$). 
This setting is commonly considered in the DA \cite{ben2010theory} and DG \cite{wang2021generalizing} literature. 
In this setting, a model trained on the data from the source domain(s) is available and we denote by $g:\X \rightarrow \Z$ the encoder of this model (we consider all layers up to the final fully connected layer as part of the encoder) and $h:\Z \rightarrow \Y$ denotes the final classifier head producing the probability distribution over the label set.

Based on this setup, our definition of transferability considers how the model trained with data from the source domain(s) performs when faced with data from potentially unseen domains. 
In contrast to previous works \cite{8803726,nguyen2020leep,huang2022frustratingly,you2021logme,tan2021otce}, proposing metrics correlated with transfer learning performance, we work in the domain generalization setting, where all the model weights are frozen i.e., neither the encoder nor the classifier is updated at test time. 
Thus, the ground truth of transferability to an unseen domain is obtained by evaluating the model's accuracy on the labeled data from the target domain formally defined below. 

\begin{definition}
\label{def:transferability}
(Transferability). Transferability of a model trained on the source domain $S$ to an unseen target domain $T$ is measured as the model's accuracy on $T$ i.e., \[\E_{(x,y) \in P_T(x,y)}[\mathrm{accuracy}(h(g(x)), y)],\] where $g:\X \rightarrow \Z$ is the pre-trained encoder and $h:\Z \rightarrow \Y$ is the pre-trained classifier producing a probability distribution over $\Y$.
\end{definition}

Measuring actual transferability as defined above requires the knowledge of the labels of the data from the unseen domains, which are not available in any practical application. 
However, the model is still expected to make correct predictions on this data. 
Moreover, the model's accuracy on the test data derived from $P_S$ may not indicate the model's performance on the data from unseen domains, especially in the presence of a distribution shift ($P_S \neq P_T$).
This makes it essential to have access to a metric that can gauge the transferability of the model at test time without requiring access to labels of the target data. 

\subsection{Background on Optimal Transport (OT)}
\label{sec:background_OT}
Optimal Transport (OT) \cite{villani2009optimal} provides a framework to compare two probability distributions in a manner consistent with their geometry. 
This has made OT a popular choice for analyzing the performance of ML models when faced with distribution shifts. 
Formally, let $\Pi(P, Q)$ be the space of joint probability distributions having $P$ and $Q$ as the marginal distributions and let $c(x_1, x_2)$ denote the dissimilarity (base distance) between two samples $x_1$ and $x_2$. 
Then, OT aims to find the coupling $\pi \in \Pi(P, Q)$ to minimize the transportation cost for moving the mass from distributions $P$ to $Q$. Mathematically, 
\begin{equation}
\label{eq:ot_distance}
\mathrm{OT}_c(P,Q) =  \inf_{\pi \in \Pi(P,Q)} \mathbb{E}_{\pi}[c(x_1, x_2)].
\end{equation}
When the cost $c(x_1, x_2) = d(x_1, x_2)^p$, where $d$ is the metric of the underlying (complete and separable) metric space, for some $p \geq 1$, then $\mathrm{OT}_c(P, Q)^{\frac{1}{p}} =: W_p(P, Q)$ denoted as the $p-$Wasserstein distance. 

Since in practice we usually only have access to finite samples, one can construct discrete empirical distributions $P=\sum_{i=1}^m a_i\delta_{x_1^i}$ and $Q=\sum_{i=1}^n b_i\delta_{y_1^i}$, where $a$ and $b$ are vectors in the probability simplex. 
In this case, the pairwise costs can be represented with a $m \times n$ cost matrix $C$ such that $C_{ij} = c(x_1^i, x_2^j)$ and the OT cost can be computed via a linear program that scales cubically in terms of the samples size. 
Since this may be prohibitive for problems with large sample sizes,  efficient solvers such as the Sinkhorn algorithm \cite{cuturi2013sinkhorn} have been proposed. 
In this work, we rely on the network simplex flow algorithm from POT~\citep{flamary2021pot} to compute the optimal coupling as detailed in the next section. 

\begin{algorithm}[t] 
\caption{Computing TETOT
} 
\label{alg:estimate_OT_distance}
\textbf{Input}: Source domain data $\mathcal{D}_S$, Target domain data $\mathcal{D}_T$, pre-trained encoder $g$ and classifier $h$, number of source samples $n$, number of target samples $m$, $\lambda$. \\
\textbf{Output}: TETOT := OT distance between source and target domain's distributions.\\
\begin{algorithmic}
\STATE{\# Select samples from source and target domains}
\STATE{Randomly sample $m$ points, $(x^i_{S},y^i_{S}) \sim \mathcal{D}_S$.}
\STATE{Randomly sample $n$ points, $(x^j_{T},y^j_{T}) \sim \mathcal{D}_T$.}
\STATE{}
\STATE{\# Compute the pairwise cost matrix based on features and \# labels}
\FOR{$i=1,\;\cdots\;,m$ and $j=1,\;\cdots\;,n$}
    \STATE{$C_{features}^{ij} = \|g(x^i_{S}) - g(x^j_{T})\|_2$}.
    \STATE{$C_{labels}^{ij} = \|y^i_{S} - h(g(x^j_{T}))\|_2$}.
\ENDFOR
\STATE{}
\STATE{\# Compute the final cost matrix}
\STATE{$C_{final} = C_{features} + \lambda C_{labels}$}.
\STATE{}
\STATE{\# Compute TETOT$:=\mathrm{OT}_c(P_S, P_T)$ by solving Eq.~\ref{eq:ot_distance} 
\begin{eqnarray*}
\small
    \min_{\pi\in\Pi(P_{S},P_T)}&&\sum_{i,j}\pi^{ij}C_{final}^{ij}\\  
    \mathrm{s.t.} &&\sum_j \pi^{ij}=\frac{1}{m}\;\forall i,\; \sum_i \pi_{ij}=\frac{1}{n}\;\forall j. 
\end{eqnarray*}
}
\STATE{return TETOT.}
\end{algorithmic}
\end{algorithm}

\subsection{Estimating transferability via TETOT}
The degradation of the performance of a model on distributions different from the ones used for training and the lack of labels of the target domain data at the time of testing makes it imperative to develop metrics that can gauge the model's performance at test time. 
Thus, we propose an OT distance-dependent metric to analyze the performance of models on unseen domains at test time.  
The efficient estimation of this metric without labels from the target domain, its intuitive meaning, and high correlation with ground truth transferability (as demonstrated in Sec.~\ref{sec:experiments}) makes it an effective way to gauge the model's performance at test time.

\begin{figure*}[t]
  \centering{
  \includegraphics[width=0.235\textwidth]{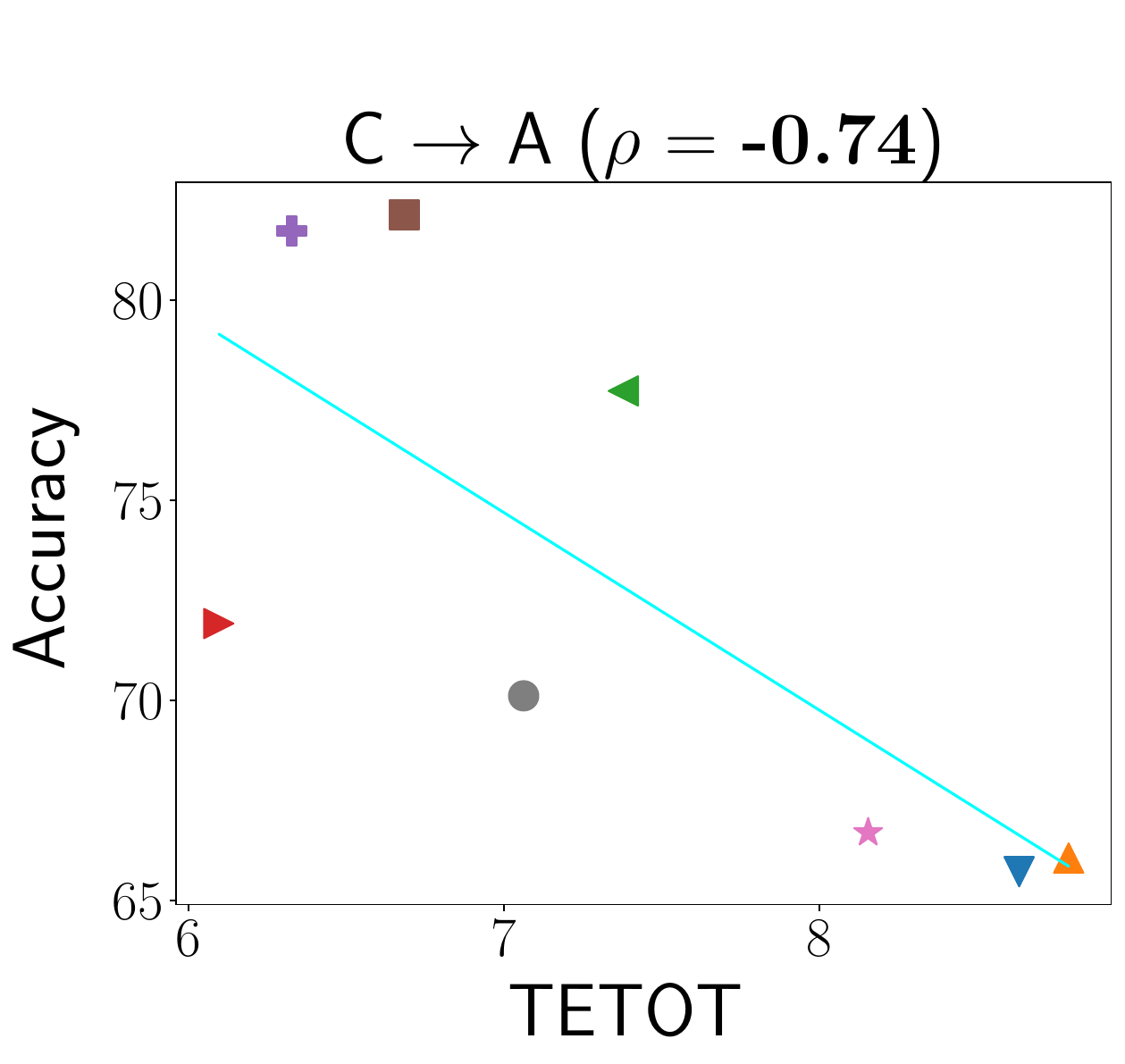}
  \includegraphics[width=0.23\textwidth]{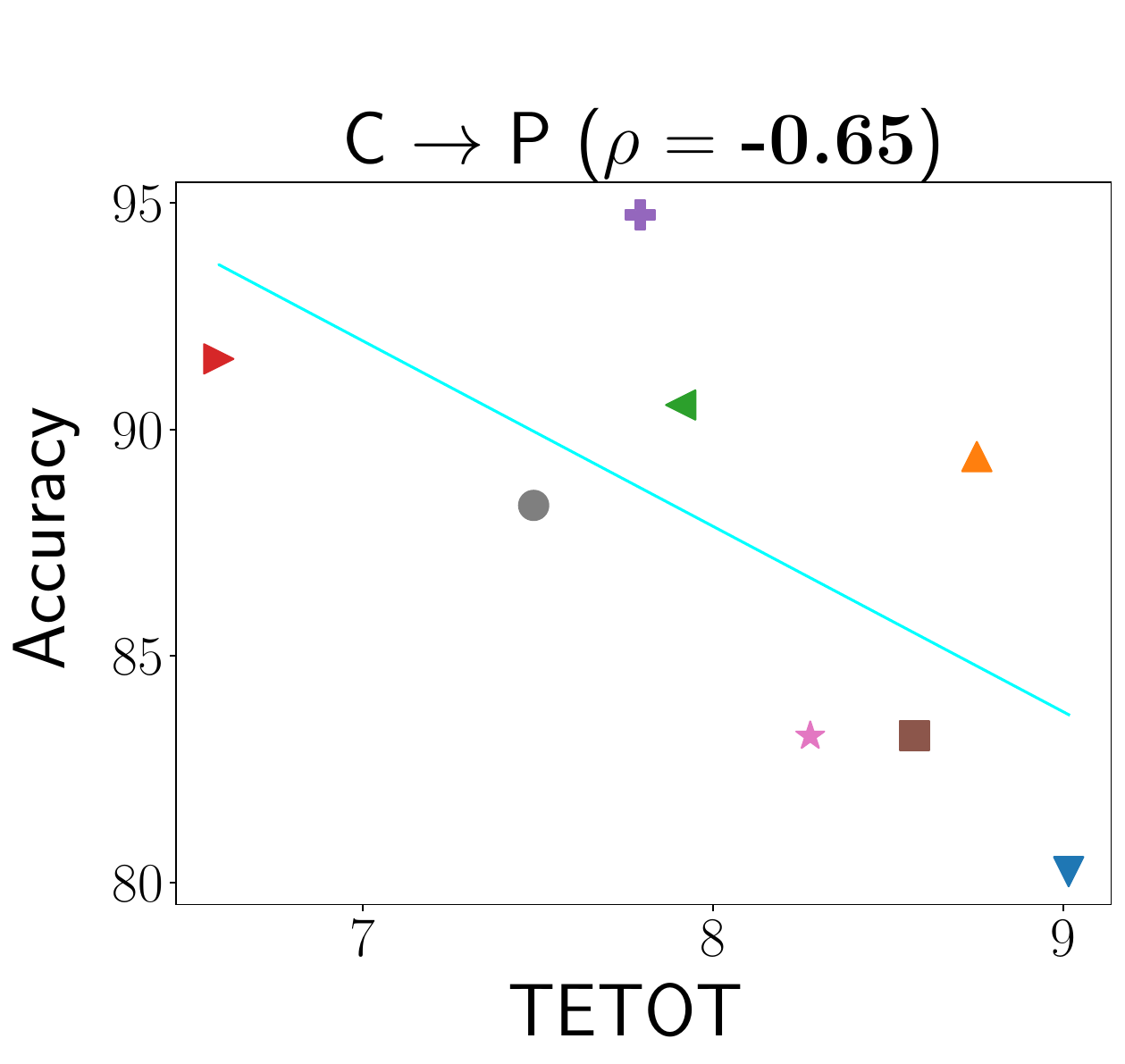}
  \includegraphics[width=0.23\textwidth]{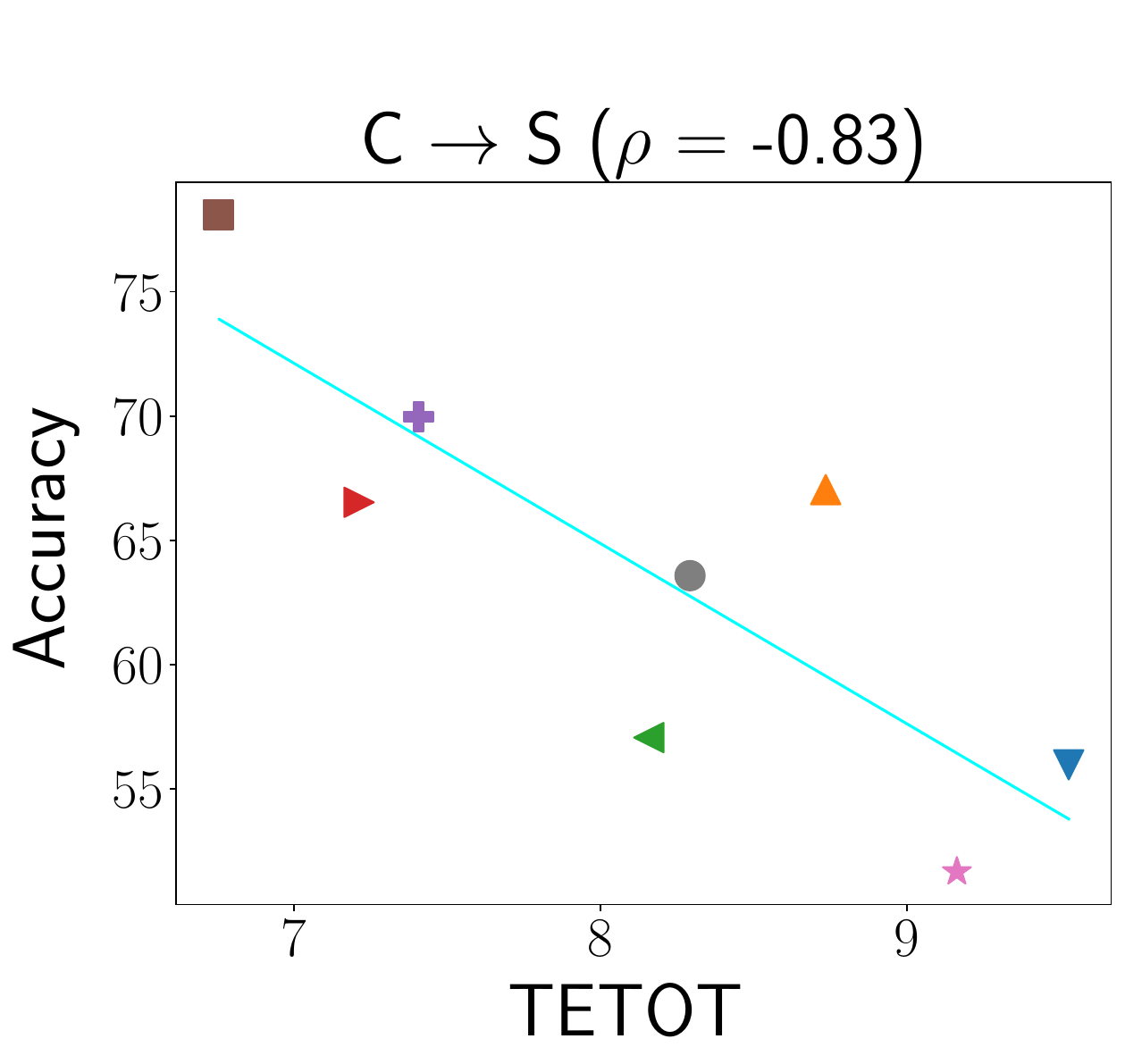}
  \includegraphics[width=0.23\textwidth]{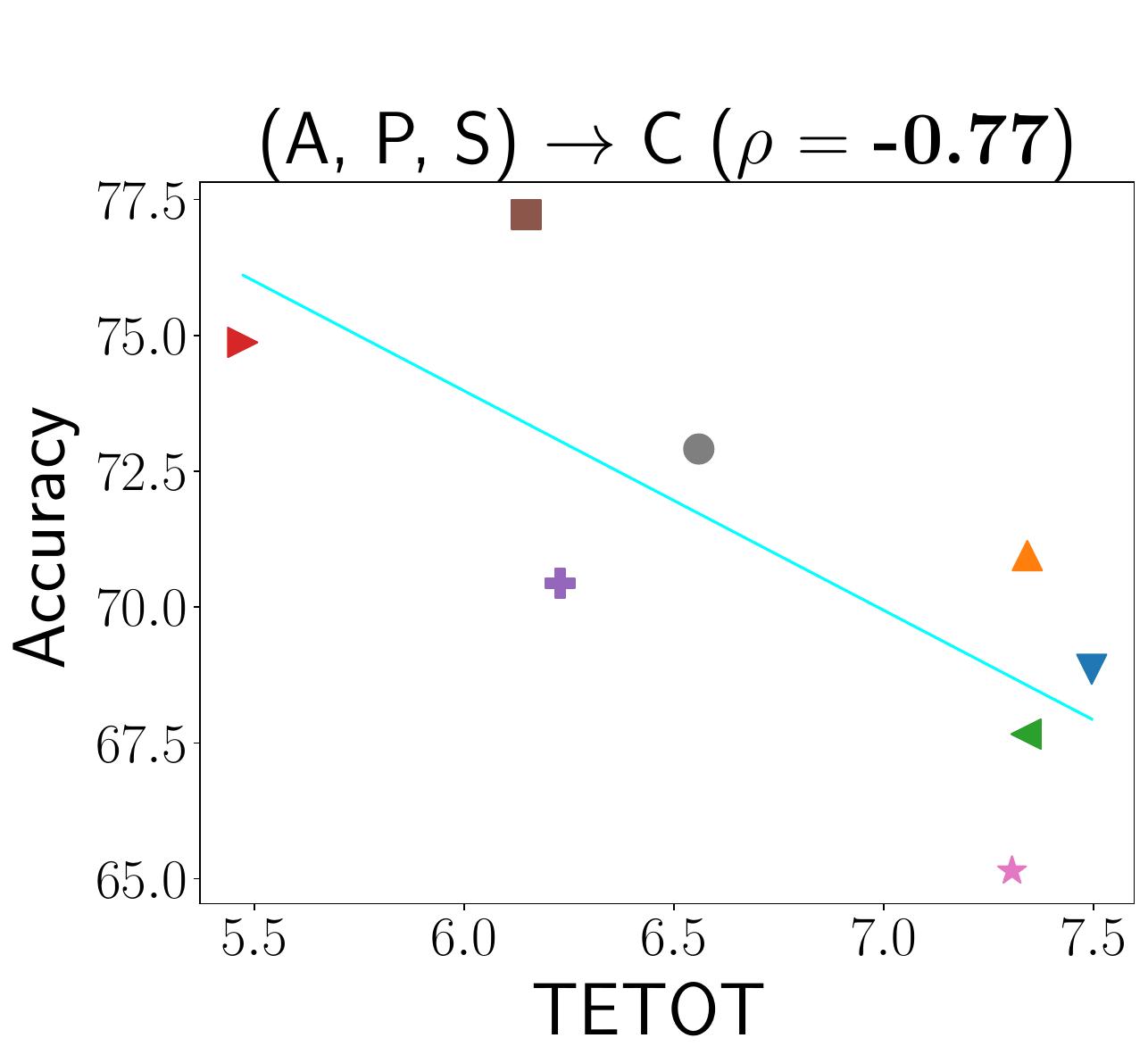}
  \includegraphics[width=0.235\textwidth]{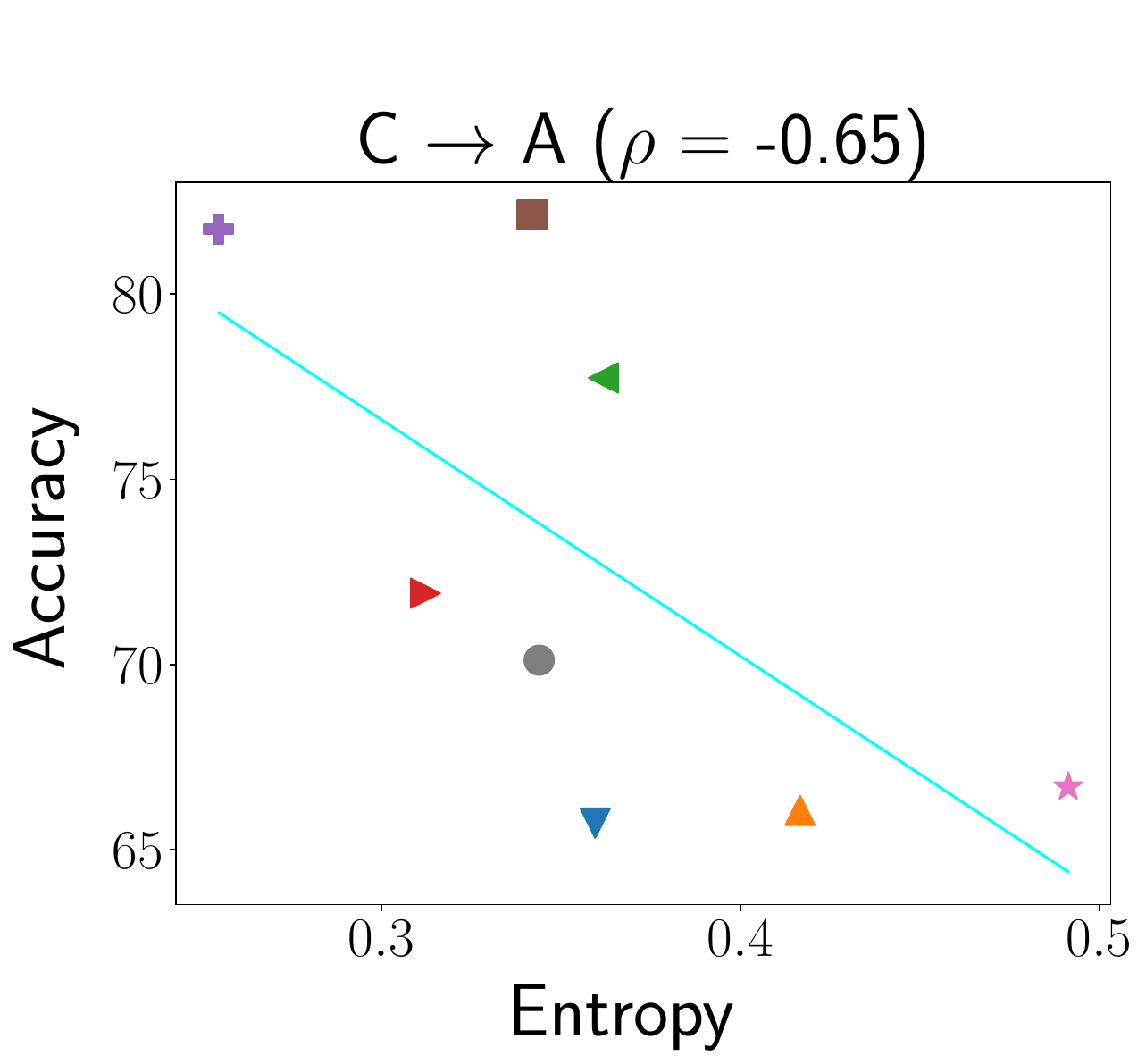}
  \includegraphics[width=0.23\textwidth]{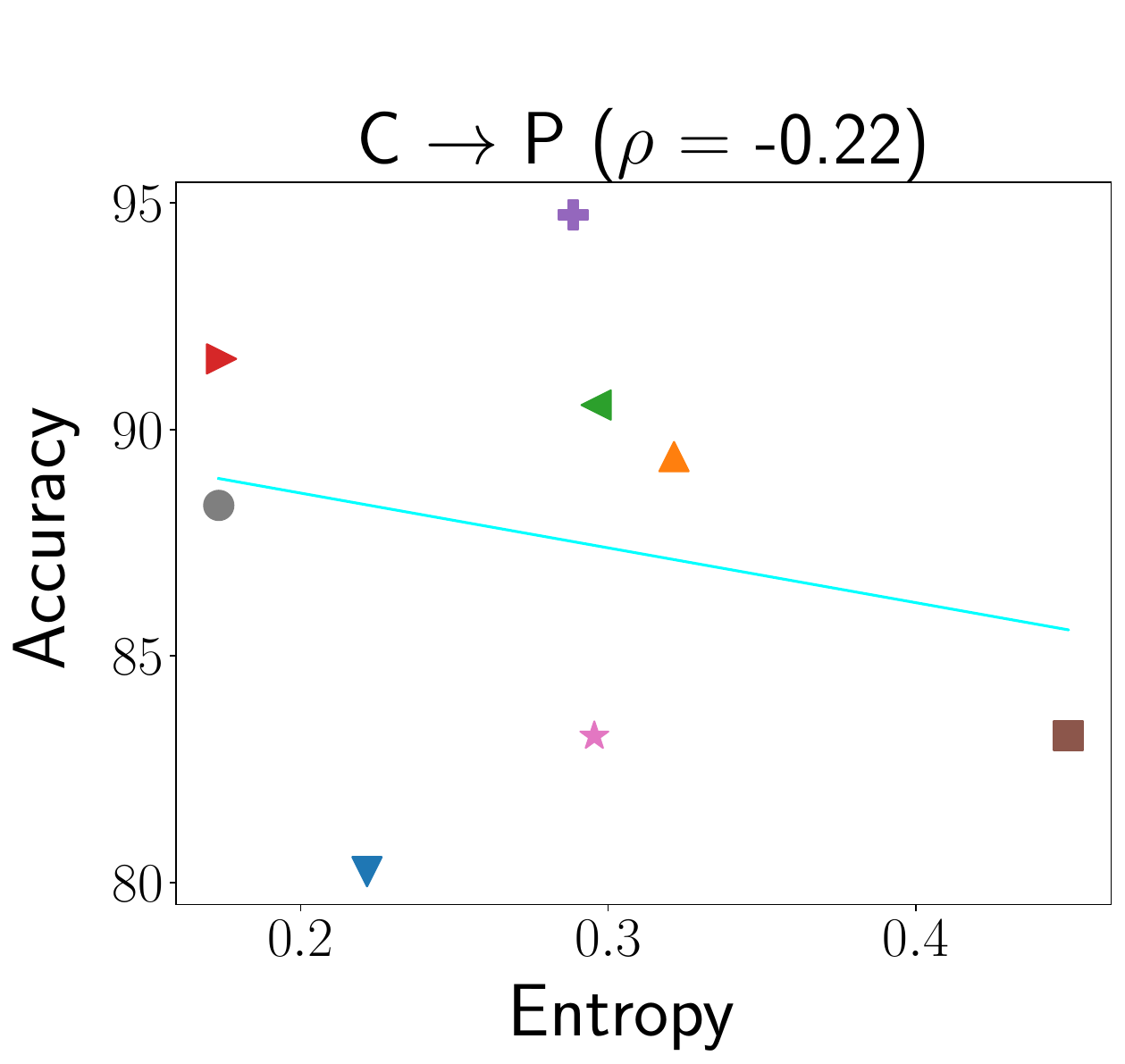}
  \includegraphics[width=0.23\textwidth]{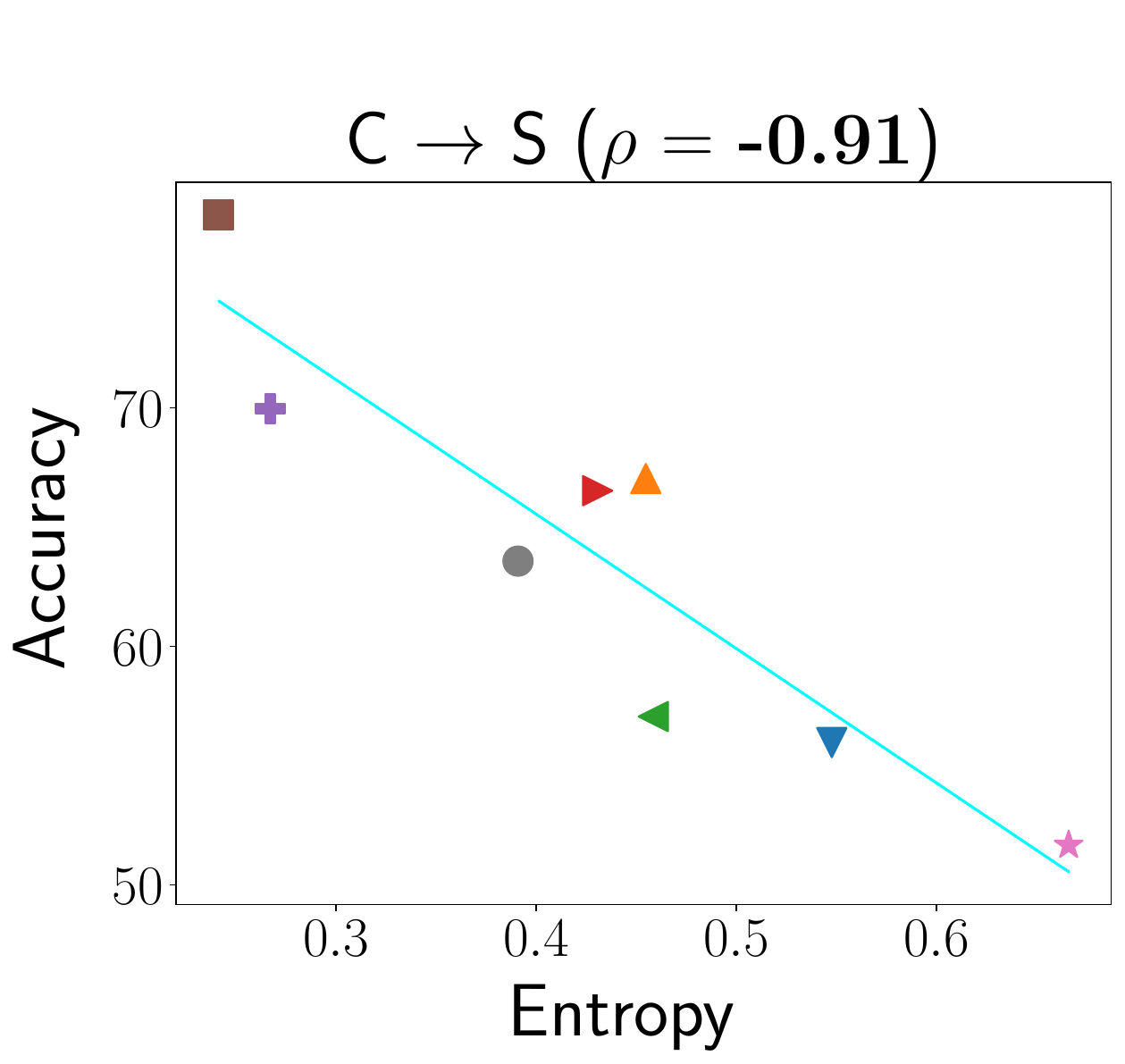}
  \includegraphics[width=0.23\textwidth]{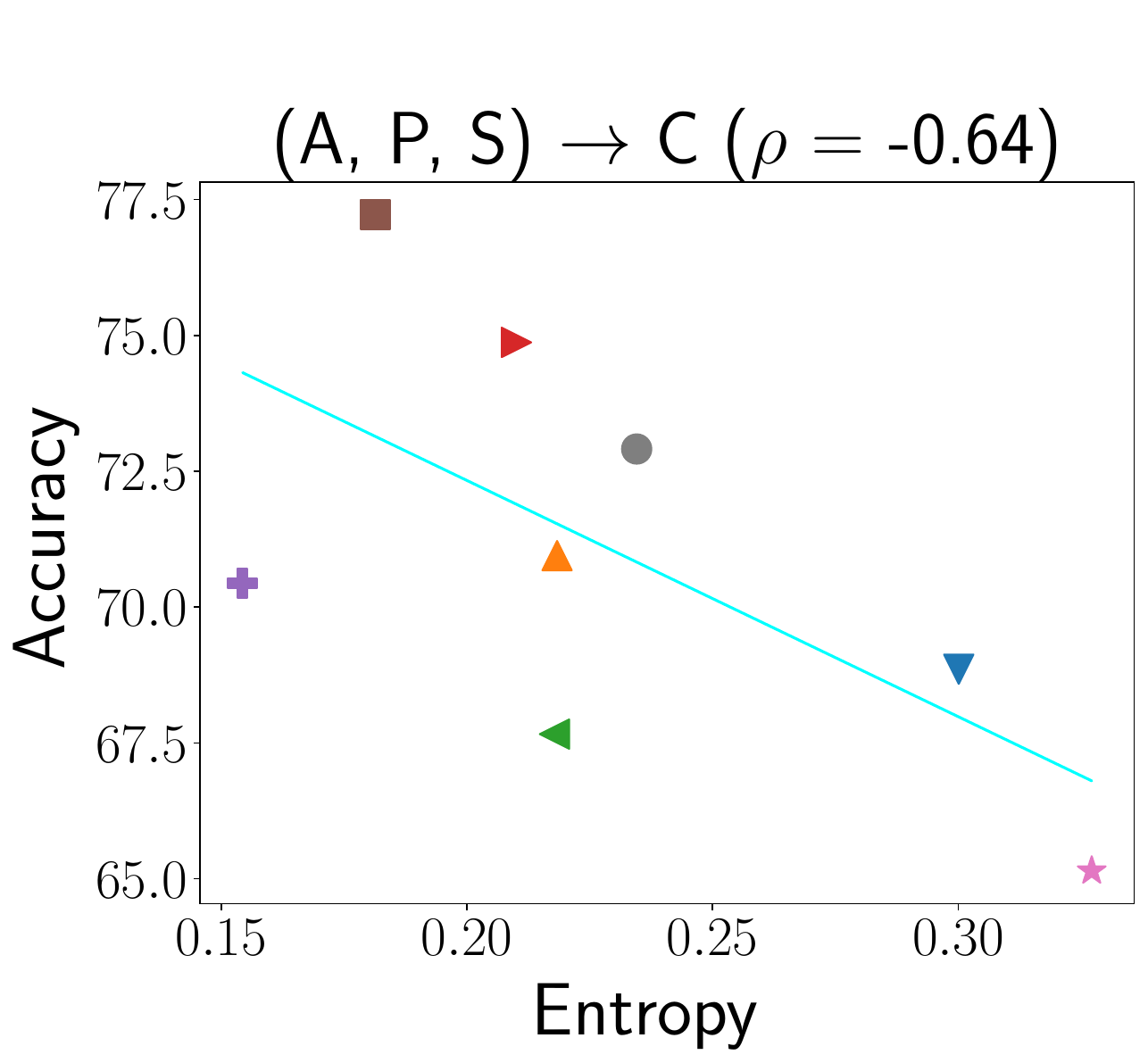}
  \includegraphics[width=0.9\textwidth]{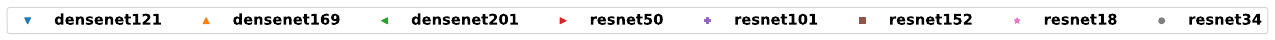}
  }
  \caption{(Best viewed in color.)
    The superiority of TETOT (top row) in achieving a high (negative) correlation ($\rho$ in the plot titles) with transferability compared to prediction entropy (bottom row) for selecting the best model architecture for making predictions on a given target domain at test time. % (with access to only unlabeled data from the target domain). 
    Models are trained using Cartoon (C) as the source domain (in the single domain setting) and Art (A), Photos (P), and Sketch (S) (in a multi-domain setting) from PACS and evaluated on various target domains.
    } 
  \label{fig:architecture_selection}
\end{figure*}

Following the line of the analytical works on understanding transferability in the DA \cite{shen2018wasserstein,ben2010theory,damodaran2018deepjdot,mehra2021understanding}, DG \cite{albuquerque2019generalizing,mehra2022do} and TL \cite{alvarez2020geometric,mehra2023analysis,tran2019transferability} settings in terms of various distributional divergence metrics, TETOT, gauges model's performance at test time based on the OT distance between the distributions of the target and the source domains.
Our base distance $c$ to be used in the computation of the OT distance consists of two parts. 
The first part focuses on measuring the dissimilarity between the features of the source and the target domains and the second focuses on measuring the difference between the labels of the two domains.
To compute the distance between the features, we follow previous works \cite{mehra2022do,damodaran2018deepjdot,volpi2018generalizing} and measure the distance between the outputs of the encoder $g$ for the source and target domains.
We define the cost $c_{features}$ using
\begin{equation}
\label{eq:feature_distance}
c_{features}(x_S, x_T) = \|g(x_S) - g(x_T)\|_2,
\end{equation}
where $x_S$ and $x_T$ denote a single sample drawn from the source and target domains, respectively. 
As shown in previous works \cite{mehra2021understanding,johansson2019support}, relying only on feature distance is not enough to explain transferability and can be improved if augmented with label information.
To this end, we propose to augment the base distance with a cost dependent on the labels of the data of the two domains. 
Since we do not have label information about the target domain, we rely on the pseudo-labels of the target domain as predicted by the source classifier. 
Many previous works have demonstrated the effectiveness of pseudo-labels in learning models for various applications including DA \cite{courty2017joint,damodaran2018deepjdot}. 
While the pseudo labels may not be aligned with the true labels of the target domain, we find that incorporating them improves the correlation of TETOT with transferability beyond only using the feature-based cost (see Sec.~\ref{sec:effect_of_lambda}).
Thus, our label cost is obtained as follows
\begin{equation}
\label{eq:label_distance}
c_{labels}(y_S, \hat{y}_T) = \|y_S - h(g(x_T))\|_2 ,
\end{equation}
where $y_S$ denotes the one-hot encoded label of the sample from the source data and $\hat{y}_T:=h(g(x_T))$ denotes the softmax output of the source classifier on the sample from the target domain.
Using Eq.~\ref{eq:feature_distance} and~\ref{eq:label_distance}, we define the base distance $c$ as

\begin{equation}
\label{eq:final_distance}
\begin{aligned}
    c((x_S, y_S), (x_T, \hat{y}_T)) = && c_{features}(x_S, x_T) \\ && + \lambda \; c_{labels}(y_S, \hat{y}_T),
\end{aligned}
\end{equation}
where $\lambda$ denotes the weight to put on the label cost. When $\lambda=0$, we recover the OT distance based on only marginal distribution matching as is common in the DA literature. 
Due to the difference between the pseudo and the true target domain labels, tuning $\lambda$ is essential. However, we find in our experiments (in Sec.~\ref{sec:effect_of_lambda}) that a value of 1 works well across various domains and applications. 
Using the base distance defined in Eq.~\ref{eq:final_distance}, we solve Eq.~\ref{eq:ot_distance} by using the network simplex flow algorithm and use the computed TETOT, to predict transferability in various applications.
The details of computing TETOT are presented in Alg.~\ref{alg:estimate_OT_distance}.

\section{Experiments}
\label{sec:experiments}
In this section, we present the evaluation results of using TETOT for predicting transferability in various practical applications as mentioned in Fig.~\ref{fig:transferability_applications}.
We show the Pearson correlation coefficient ($\rho$) between TETOT and ground truth transferability in our evaluation. A high negative correlation ($\rho$) implies higher transferability between the source and target domains. 
Along with applications mentioned in Fig.~\ref{fig:transferability_applications}, we also show how to estimate transferability using only statistics from the source data, which may be required in scenarios where source data is costly to store or private and hence cannot be accessed at test time. 
Following this, we show the effect of the number of samples used for estimating the metric and the sensitivity of the metric to the choice of $\lambda$ in Eq.~\ref{eq:final_distance}.

\begin{table}[t]
  \begin{center}
    \captionof{table}{TETOT achieves a higher (negative) correlation to transferability than entropy on the architecture selection problem on PACS and VLCS datasets.
      \label{table:architecture_selection}
    }
    \resizebox{0.25\textwidth}{!}{
    \begin{tabular}{|c|cc|}

      \hline
     Dataset & Entropy & TETOT  \\
     \hline
     PACS & -0.40 & {\bf -0.62} \\
     VLCS & -0.29 & {\bf -0.40} \\
     \hline
     Average & -0.35 & {\bf -0.51} \\
      \hline

    \end{tabular}
    }
  \end{center}

\end{table}

\begin{table}[t]
  \begin{center}
    \captionof{table}{TETOT achieves a superior correlation compared to entropy for selecting the source domain that achieves the highest transferability to a target domain.
      \label{table:source_selection}
    }
    \resizebox{0.25\textwidth}{!}{
    \begin{tabular}{|c|cc|}

      \hline
     Dataset & Entropy & TETOT  \\
     \hline
     PACS & -0.47 & {\bf -0.94} \\
     VLCS & -0.58 & {\bf -0.92} \\
     \hline
     Average & -0.53 & {\bf -0.93} \\
      \hline
    \end{tabular}
    }
  \end{center}
\end{table}

\begin{table}[t]
  \begin{center}
    \captionof{table}{TETOT is significantly better at predicting transferability of a model to unseen domains (achieves a higher (negative) correlation to transferability) compared to entropy on PACS and VLCS datasets.
      \label{table:predicting_unseen_domain_acc}
    }
    \resizebox{0.25\textwidth}{!}{
    \begin{tabular}{|c|cc|}

      \hline
     Dataset & Entropy & TETOT  \\
     \hline
     PACS & -0.39 & {\bf -0.93} \\
     VLCS & -0.34 & {\bf -0.80} \\
     \hline
     Average & -0.36 & {\bf -0.86} \\
      \hline

    \end{tabular}
    }
  \end{center}

\end{table}

Our evaluation includes two popular benchmark datasets, PACS \cite{li2017deeper} and VLCS \cite{fang2013unbiased} with four domains each.
PACS \cite{li2017deeper} consists of 9991 images from Art, Cartoons, Photos, and Sketches across 7 different classes and 
VLCS \cite{fang2013unbiased} consists of 10729 images from Caltech101, LabelMe, SUN09,
PASCAL VOC 2007 across 5 different classes.
We also present an evaluation of corrupted versions of these datasets to mimic different unseen distributions. 
We use common corruptions \cite{hendrycks2018benchmarking} including \texttt{\{brightness, contrast, spatter, saturate, elastic transform, gaussian blur, defocus blur, zoom blur, gaussian noise, shot noise, impulse noise, speckle noise\}} with 5 different severity levels in our evaluation. 

For our experiments, we fine-tune models pre-trained on the Imagenet dataset with ERM using labeled data from the source domain with an addition of a 128-dimensional bottleneck layer followed by a linear fully connected layer, with 7 units for the PACS and 5 units for the VLCS dataset, for classification. 
To make OT distance from different models/datasets comparable we normalize the output of the trained encoder separately for source and target representations before computing $C_{feature}$ in all experiments except for those in Sec.~\ref{sec:source_statistics}.

We compare the correlation between transferability and our metric on various domains, with a popular target data-only dependent metric of prediction entropy (computed as 
\(\frac{1}{n}\sum_{x \in \mathcal{D}_T}H(h(g(x)),
\mathrm{where} \; H(\hat{y}) = -\sum_{c \in \Y}p(\hat{y}_c)\log(p(\hat{y}_c))\). 
This metric has previously been shown to achieve high correlation with transferability and thus used in various test-time adaptation \cite{wang2020tent,mummadi2021test,thopalli2023domain,liang2020we} works to adapt the model to distribution shifts. 
While the prediction entropy produces a high correlation with transferability, TETOT which uses both source and target data achieves a significantly better correlation across various applications and can be computed merely in $\approx$2 seconds on the PACS dataset on our hardware. Our code can be found at \url{https://github.com/akshaymehra24/TETOT}.

\begin{figure*}[t]
  \centering{
  \includegraphics[width=0.23\textwidth]{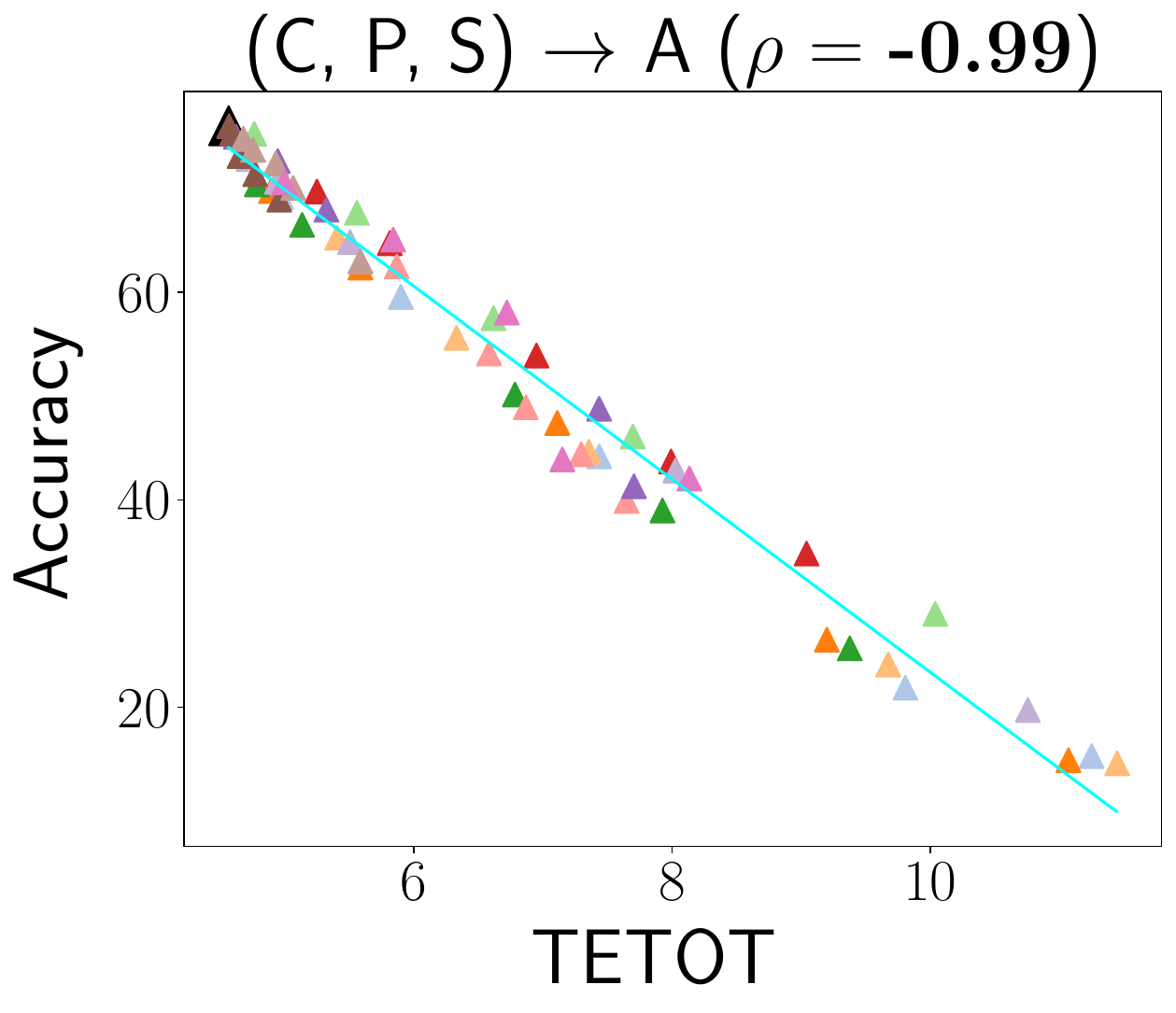}
  \includegraphics[width=0.23\textwidth]{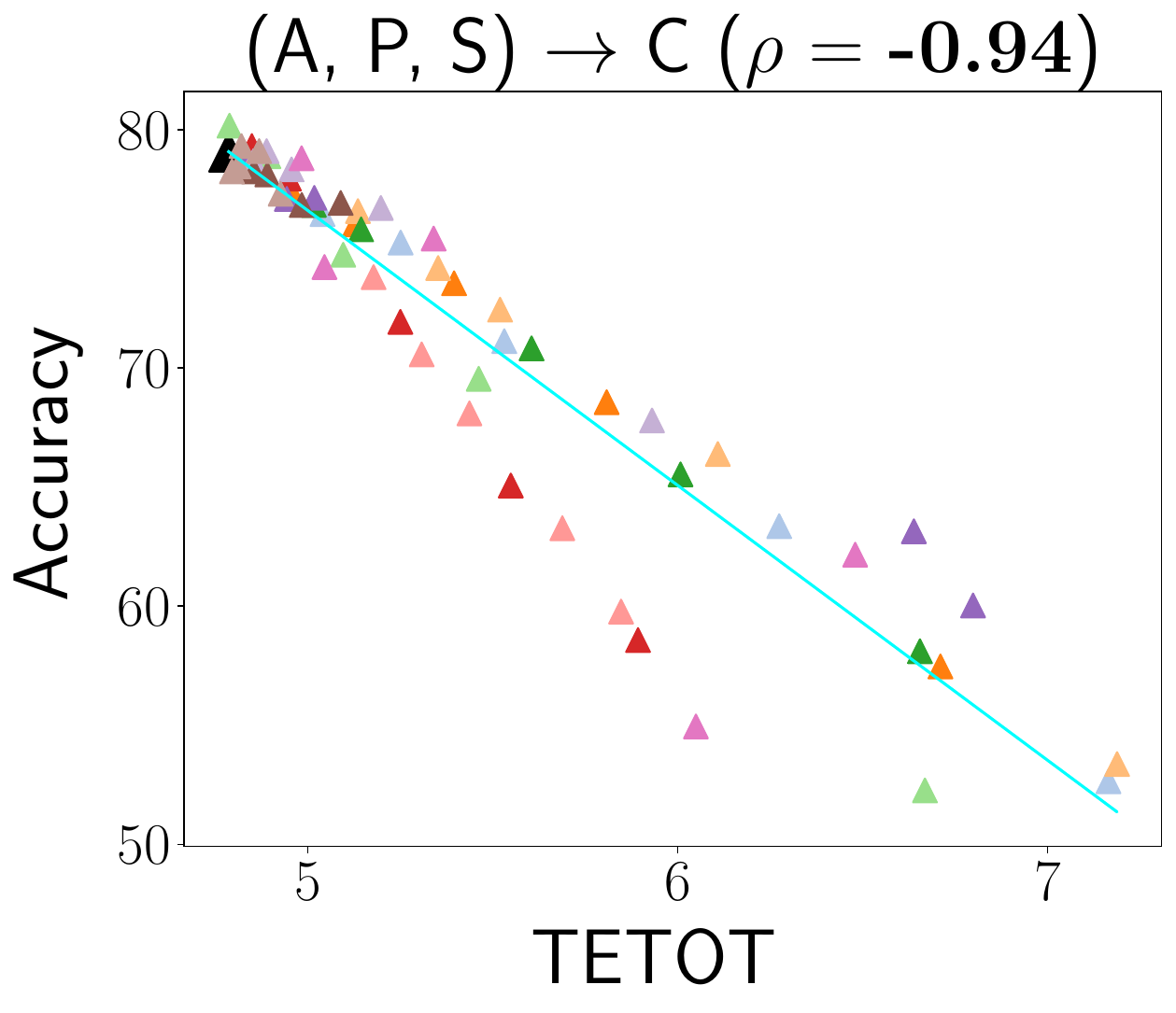}
  \includegraphics[width=0.23\textwidth]{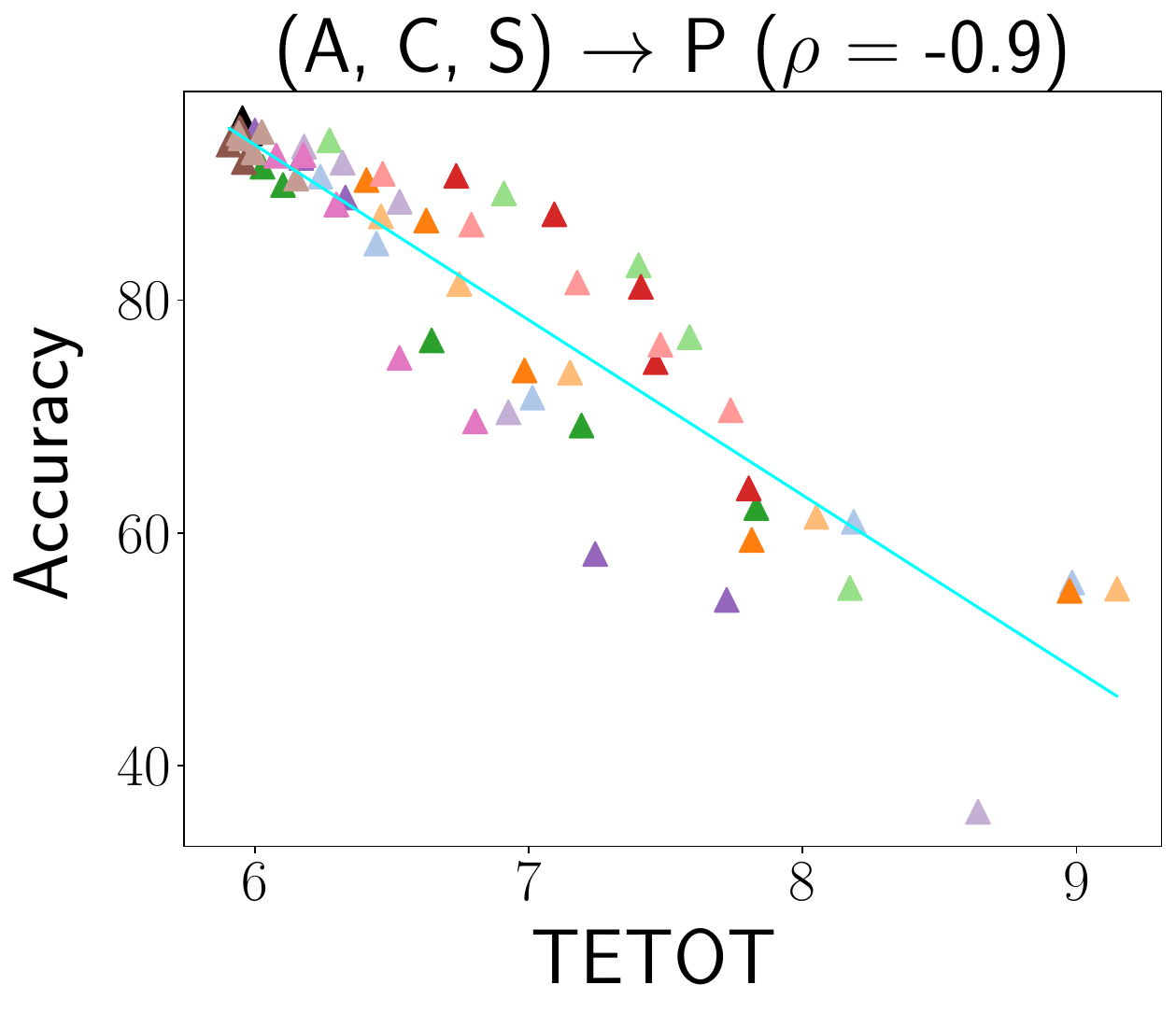}
  \includegraphics[width=0.23\textwidth]{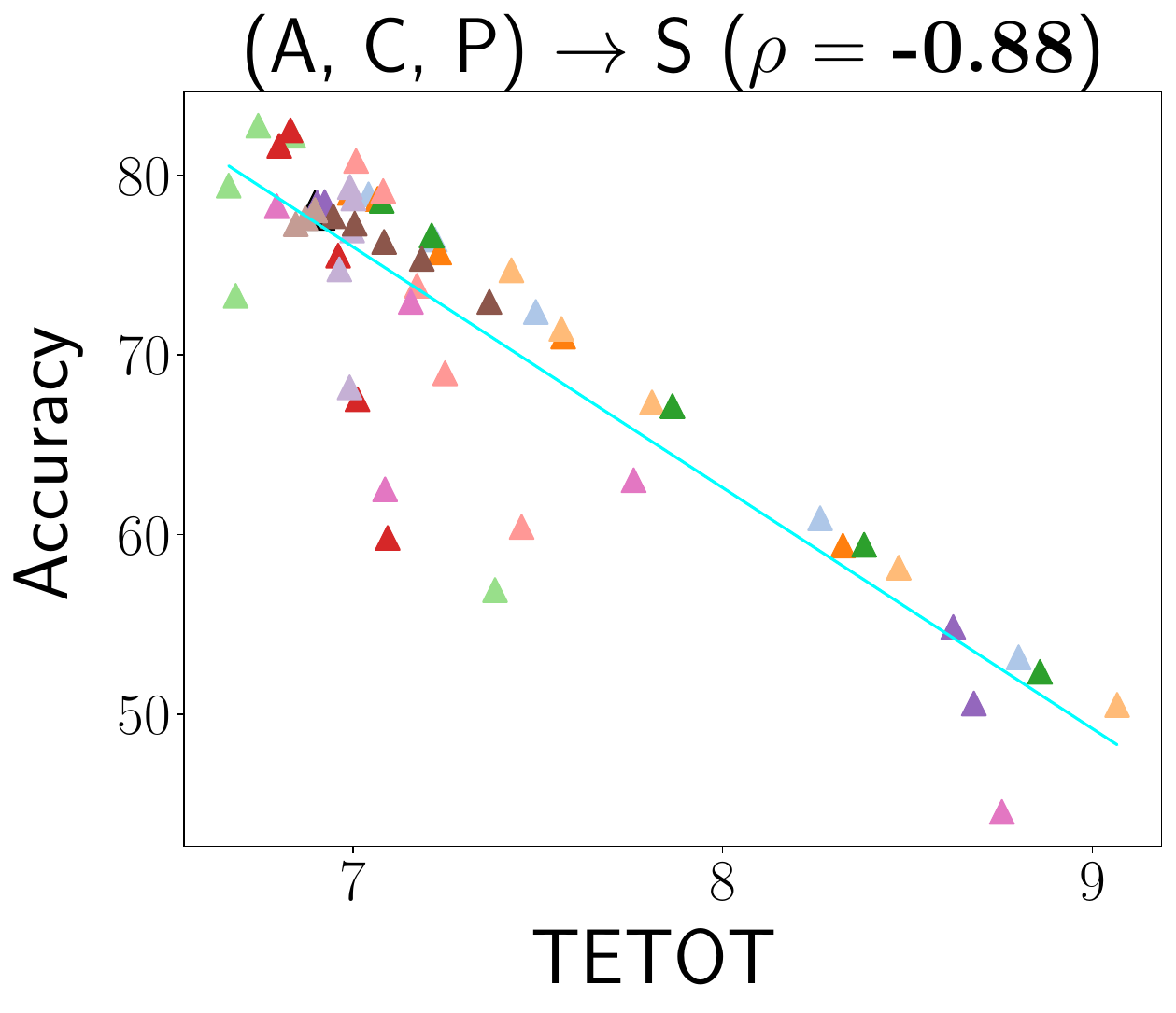}
  \includegraphics[width=0.23\textwidth]{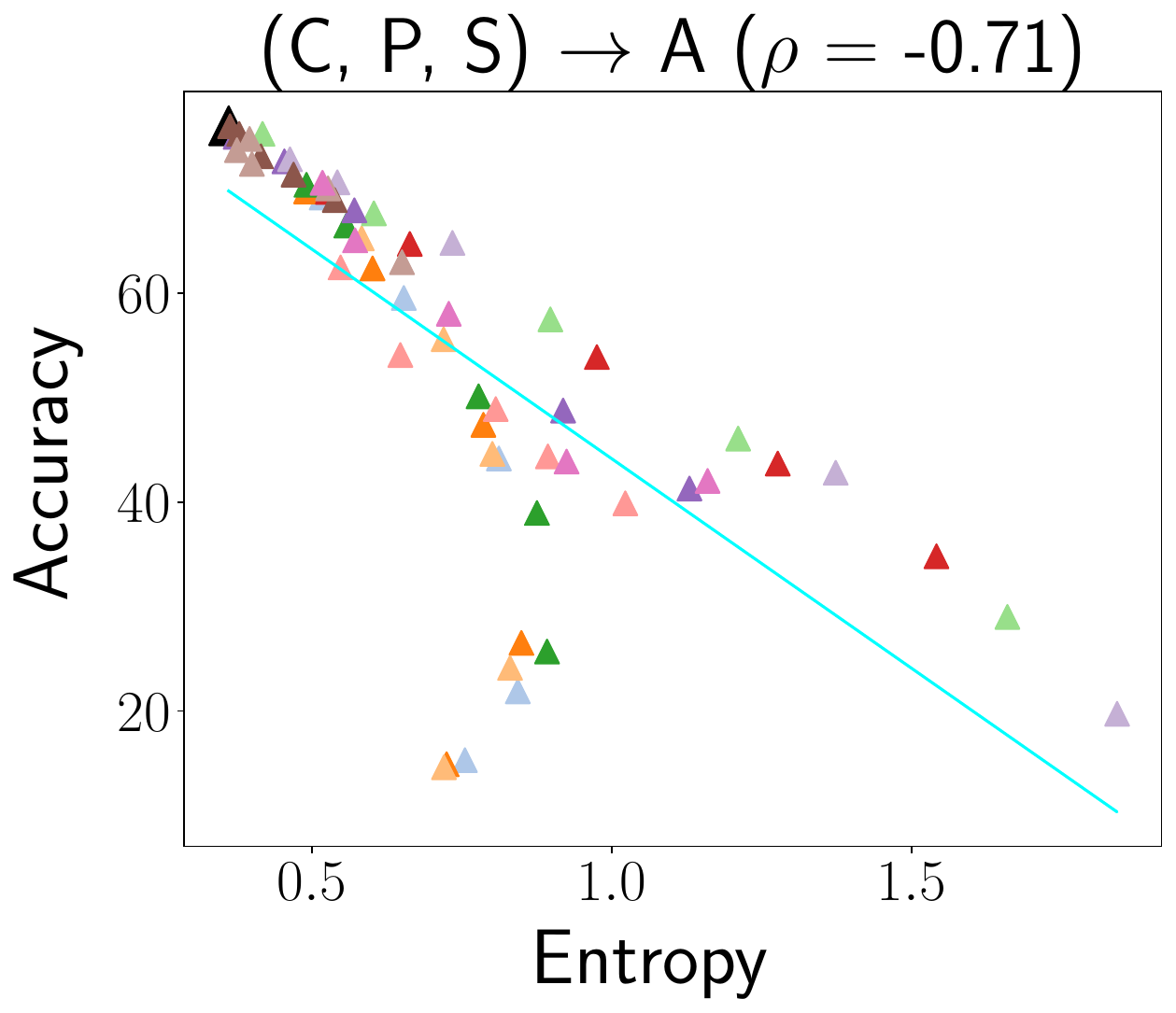}
  \includegraphics[width=0.23\textwidth]{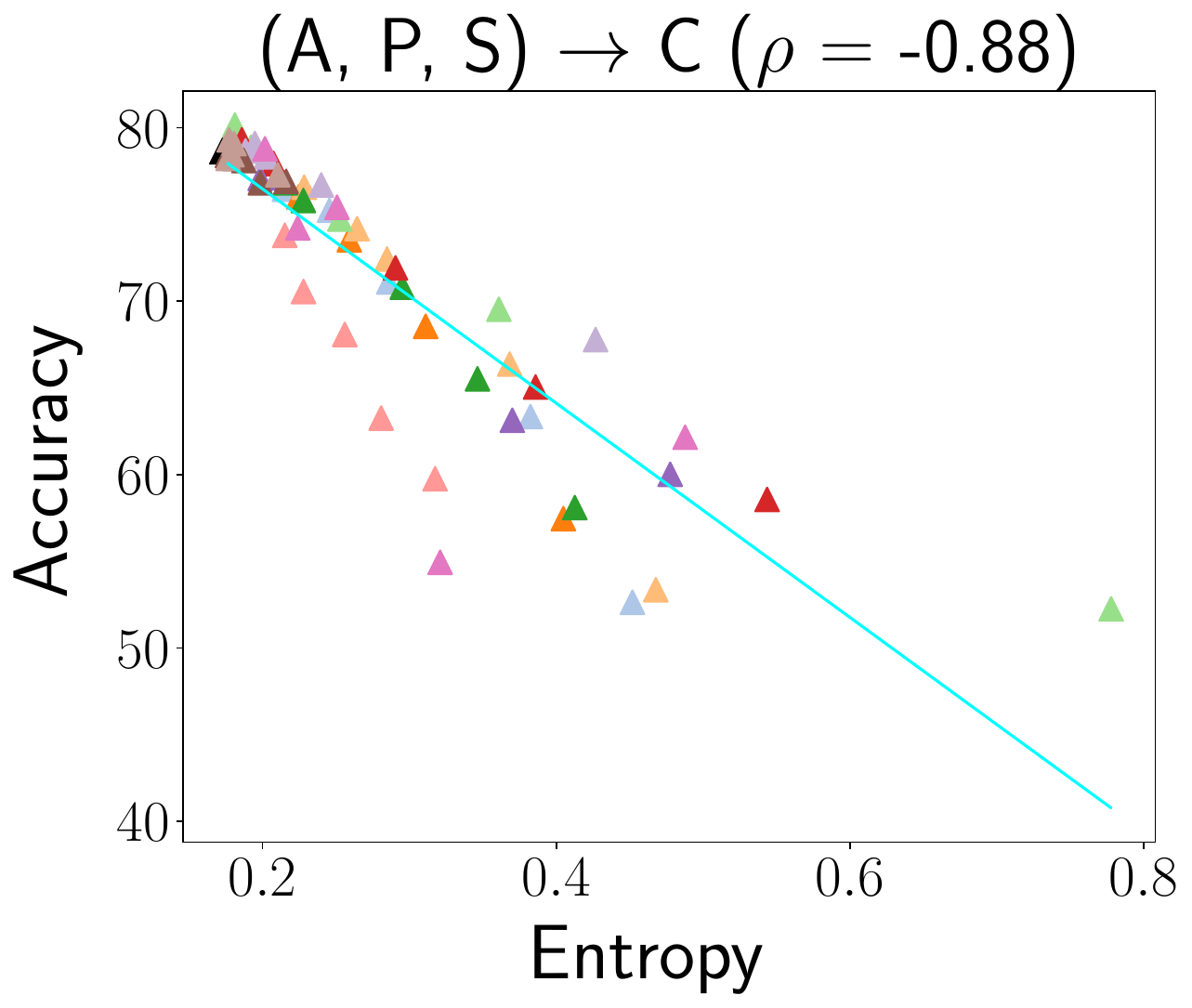}
  \includegraphics[width=0.23\textwidth]{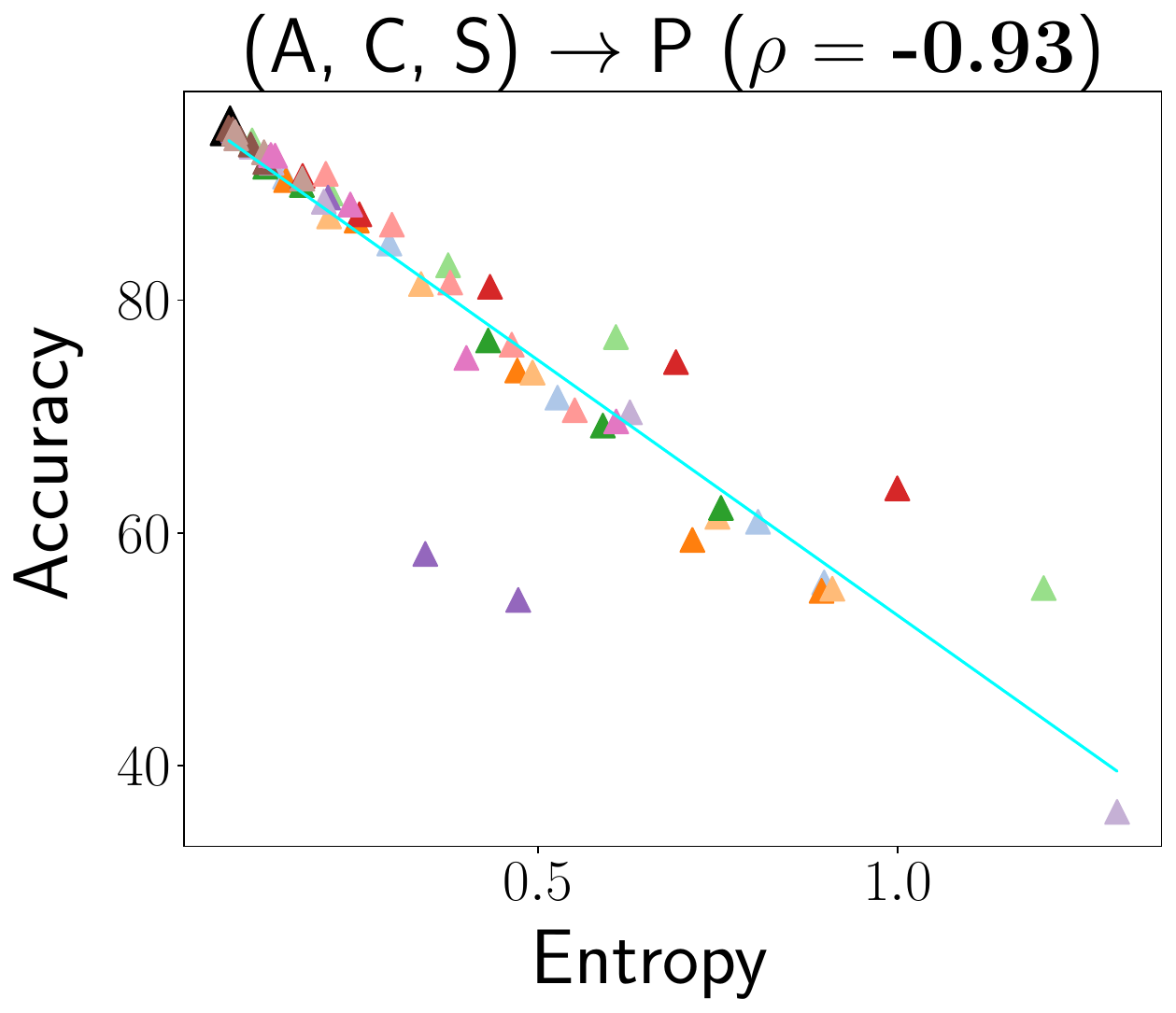}
  \includegraphics[width=0.23\textwidth]{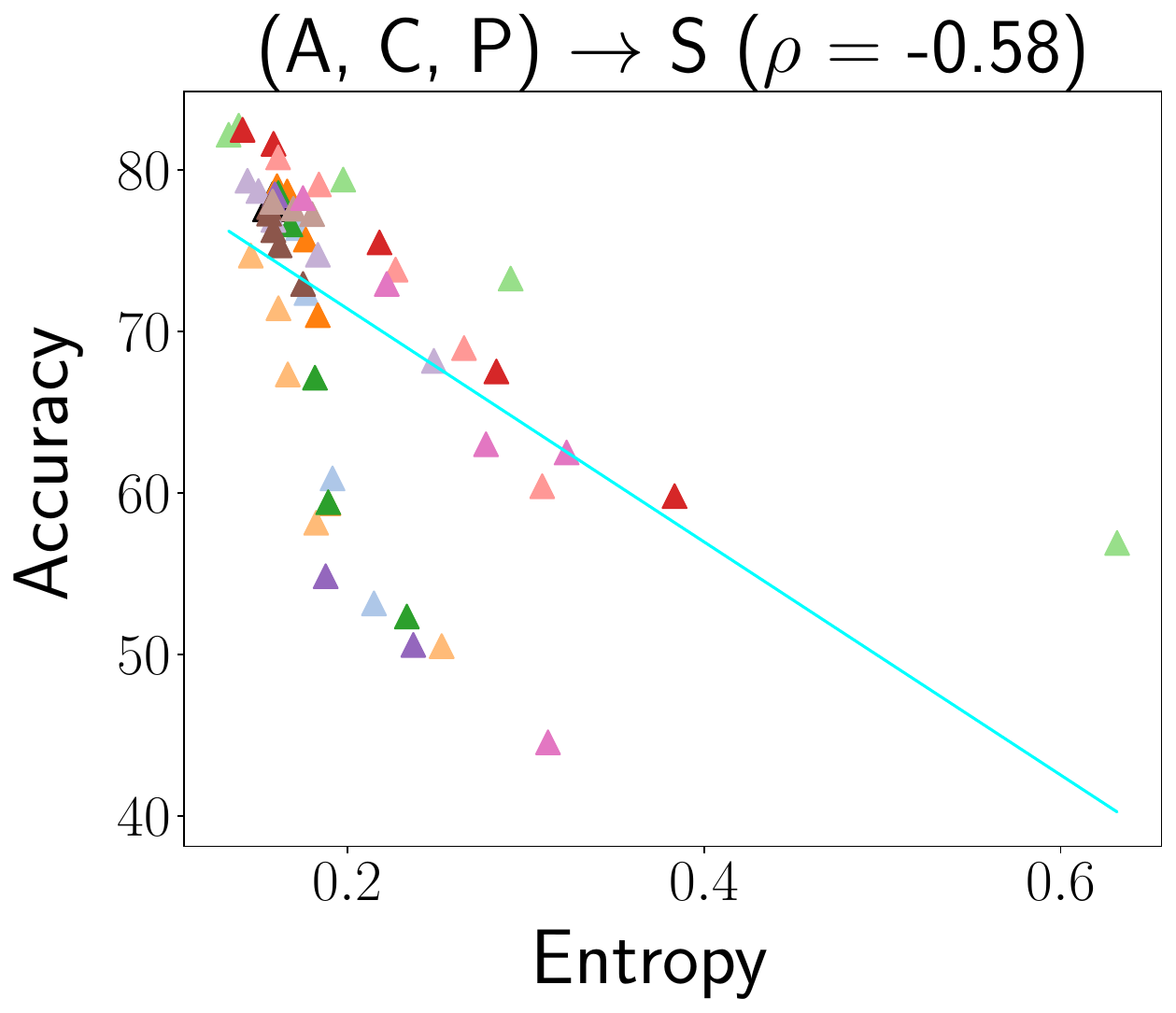}
  \includegraphics[width=0.95\textwidth]{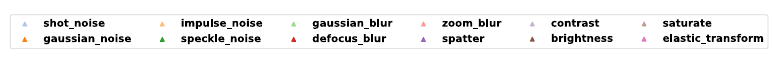}
  }
  \caption{(Best viewed in color.)
    The superiority of TETOT (top row) compared to prediction entropy (bottom row) in achieving a high (negative) correlation ($\rho$ in the plot titles) with transferability on unseen domains encountered at test time. 
    Models are trained using multiple source domains and evaluated on an unseen target domain from the PACS dataset. (The black triangle denotes the original data of the target domain whereas the colored triangles denote the target domain data corrupted by different corruptions and severity levels)
    } 
  \label{fig:unseen_domain_accuracy_prediction}
\end{figure*}

\subsection{Architecture selection for a target domain}
\label{sec:architecture_selection}
In this section, we show how TETOT can be used for selecting the model architecture that leads to the highest transferability to an unseen target domain at test time. 
For this experiment, we used eight model architectures including ResNet18, ResNet34, ResNet50, ResNet101, ResNet152, DenseNet121, DenseNet169, and DenseNet201.
We consider models trained on PACS and VLCS in both single and multi-domain settings. 
For the single-domain setting, the models are trained using only one of the four domains from these datasets, whereas in the multi-domain setting, the models are trained using three out of the four domains. 
Then, given a target domain (e.g., Art from PACS), we compute the correlation between the transferability and TETOT of all eight architectures trained on a particular source domain(s) (e.g., Cartoon from PACS). 
The results in Fig.~\ref{fig:architecture_selection} show that TETOT achieves a high correlation with transferability on this problem, significantly better than using prediction entropy. 
Moreover, our results in Table~\ref{table:architecture_selection}, averaged over all 32 target domains (12 for single domain and 4 for multi-domain setting for each dataset) in PACS and VLCS for models trained in both single and multi-domain settings show the high correlation achieved by TETOT with transferability on this problem and show the superiority of TETOT compared to prediction entropy.
This demonstrates that TETOT can be used to select the model architecture to use for making the predictions at the test time with just access to unlabeled data from the target domain.

\subsection{Source dataset selection for a target domain}
Here we present the results of using TETOT to select the model trained with the best source domain for a given target domain. 
For this experiment, we fix the model architecture to ResNet50 and train it in both single and multi-domain settings on PACS and VLCS datasets.
Then for a given target domain, e.g., Art from PACS, we aim to use TETOT to find the model that achieves the highest transferability to the target domain, among the models trained on Cartoon, Photos, Sketch, or a combination of the three. %, such that the model has the highest transferability to the target domain.
The averaged results over 8 different target domains (4 from PACS and 4 from VLCS) in Table~\ref{table:source_selection}, show that TETOT significantly outperforms prediction entropy and achieves almost a perfect negative correlation with transferability on this problem. 
This highlights the effectiveness of TETOT in selecting the best source domain that provides the highest transferability to a target domain.

\subsection{Assessing transferability to unseen domains}
In this section, we evaluate the performance of TETOT in predicting the transferability of a given model to data from unseen domains. 
For this experiment, we use ResNet50 models trained with different source domains in both single and multi-domain settings on PACS and VLCS datasets. 
Using a fixed model, we evaluate how TETOT computed using only the unlabeled data from a target domain correlates with the accuracy of the model on this target domain.
Results in Fig.~\ref{fig:unseen_domain_accuracy_prediction}, show that TETOT is highly (negatively) correlated with the accuracy of the model on various target domains. 
To generate a wide array of unseen target domains, we add corruptions (with 5 different severity levels) to the original target domain similar to those used in \cite{hendrycks2019robustness,mehra2024fly}.
A high correlation of TETOT with transferability (better than entropy as shown in Table~\ref{table:predicting_unseen_domain_acc}) allows TETOT to be used at test time to get insights into the performance of the model on unseen domains.
Specifically, suppose the TETOT metric is high compared to the TETOT metric of the original test set (which can be saved as a reference during the model training stage). 
In that case, likely, the model is not producing good predictions on the unseen domain. 
Based on this information, test-time adaptation procedures such as TENT \cite{wang2020tent} can be invoked and the model can be adapted to the distribution of the unseen domain.
Thus, TETOT is a useful metric for gauging model performance at test time.

\subsection{Effect of sample size on TETOT}
\begin{figure}[tb]
  \centering{
  \includegraphics[width=0.30\textwidth]{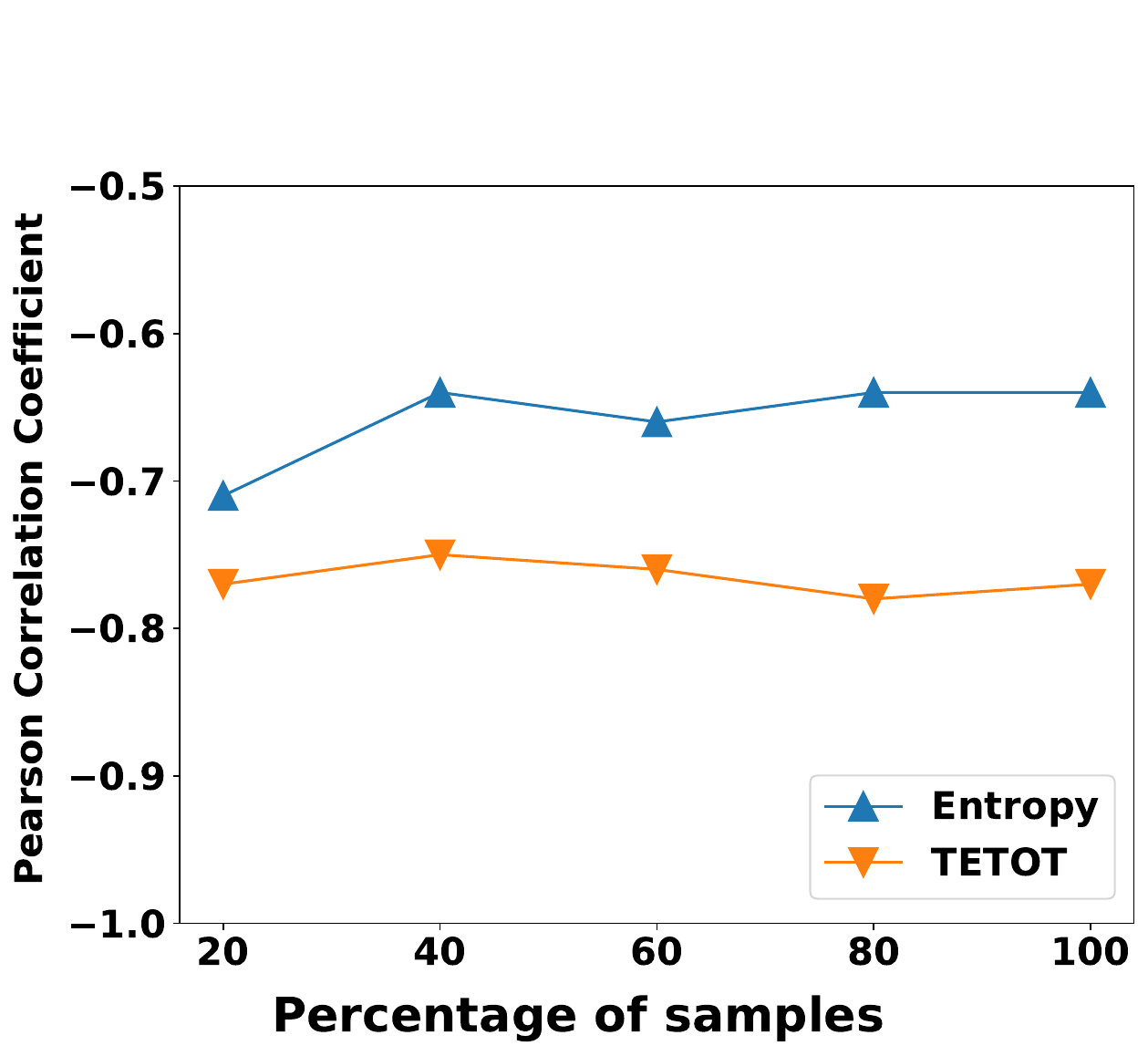}
  }
  \caption{
    The Pearson correlation coefficient between transferability and TETOT remains high and better than entropy (a higher negative correlation is better i.e., a smaller number is better) for different sample sizes of the source and target domains. 
    } 
  \label{fig:effect_of_sample_size}
\end{figure}

In this section, we show that TETOT achieves a high correlation with transferability even by using a portion of the data from the source and target domains.
For this experiment, we present the correlation between transferability and TETOT on the architecture selection problem for the case when the target domain is fixed to Cartoon and the eight different architectures (considered in Sec.~\ref{sec:architecture_selection}) are trained using the remaining three domains from the PACS dataset.
The results in Fig.~\ref{fig:effect_of_sample_size}, show that the Pearson correlation coefficient between TETOT and transferability remains consistently high for different proportions of the source and target domain data on this problem.
%Mention that entropy correlations get slightly worse with more data if we can explain it.

\subsection{Estimating transferability without source data}
\label{sec:source_statistics}
In practical scenarios, it may be difficult to access the source dataset (due to privacy constraints or memory limitations of devices used for model deployment) used for training the models, making transferability estimation challenging. 
Using the flexibility of the OT framework, we propose a metric that can be estimated with just the statistics (mean and covariance) of the source data avoiding the need to access the entire source dataset.
Specifically, we use the following cost $c=c_{features}^2$ in Eq.~\ref{eq:final_distance}. 
Using this cost, the square root of the optimal transport distance computed using Eq.~\ref{eq:ot_distance} is the same as the 2-Wasserstein distance as mentioned in Sec.~\ref{sec:background_OT}.
To simplify the computation further, we assume that distributions of the source and target domains follow the Normal distribution. 
This allows us to estimate the 2-Wasserstein distance in a closed form using only on the statistics of the source and target domains.
Let $\hat{\mu}_S = \frac{1}{m}\sum_{x \in \mathcal{D}_S}g(x)$, $\hat{\Sigma}_S = \frac{1}{m}\sum_{x \in \mathcal{D}_S}(g(x)-\hat{\mu}_S)^T(g(x)-\hat{\mu}_S)$ and let $\hat{\mu}_T$, $\hat{\Sigma}_S$ be defined in a similar manner, then $\mathrm{OT}(P_S, P_T)$ is 
\begin{equation}
\begin{aligned}
    W_2^2(P_S,P_T)= \|\hat{\mu}_S - \hat{\mu}_T\|_2^2 
    + \mathrm{tr}(\hat{\Sigma}_S + \hat{\Sigma}_T - 2(\hat{\Sigma}_S^{\frac{1}{2}}\hat{\Sigma}_T\hat{\Sigma}_S^{\frac{1}{2}}).
\end{aligned}
\end{equation}
We refer to this as approximate TETOT (TETOT-approx).
Our results in Table~\ref{table:architecture_selection_using_FID} show that this metric also achieves a high correlation with transferability while being slightly worse than TETOT on the problem of architecture selection on the PACS dataset.
Similar to the previous section, the target domain is fixed to Cartoon from PACS, and the transferability of eight architectures trained on the other three domains is estimated via TETOT.

\subsection{Effect of different $\lambda$ in Eq.~\ref{eq:final_distance} for TETOT}
\label{sec:effect_of_lambda}
In this section, we show how $\lambda$ the coefficient of the label cost in Eq.~\ref{eq:final_distance} influences the correlation of TETOT and transferability.
We test $\lambda \in \{$0, 1, 1E2, 1E4$\}$, where $\lambda=$0 corresponds to only using the marginal feature distribution for computing the distance between the distributions of the source and target domains. 
Our results in Table~\ref{table:effect_of_lambda} show that $\lambda=$1 achieves the highest correlation on both PACS and VLCS datasets for the problem of architecture selection using models trained in the multi-domain setup, where three domains are used for training the eight architectures (in Sec.~\ref{sec:architecture_selection}) and the remaining domain is used as the target domain for evaluation. 
We also see that $\lambda=$1 outperforms the case of marginal matching (with $\lambda=$0), suggesting that incorporating label distance is useful for improving the correlation between TETOT and transferability.
However, since we only have access to pseudo labels for the data from the target domain which may be incorrect, emphasizing the label cost, to force label-wise matching, degrades the correlation. 
Thus a small label cost with $\lambda=$1 yields the highest correlation (-0.51) and performs competitively to the correlation obtained using true labels for the target domain (-0.55).

\begin{table}
  \begin{center}
    \captionof{table}{Effective transferability estimation with TETOT-approx $:= W_2^2(P_S,P_T)$ using only statistics from the two domains. 
      \label{table:architecture_selection_using_FID}
    }
    \resizebox{0.34\textwidth}{!}{
    \begin{tabular}{|c|c|}
      \hline
     Metric & Pearson Corr. Coeff. \\
     \hline
     TETOT-approx & -0.60\\
     %Entropy & -0.60\\
     TETOT &  -0.75\\
     \hline
    \end{tabular}
    }
  \end{center}

\end{table}

\begin{table}
  \begin{center}
    \captionof{table}{Effect of using different values of $\lambda$ in the base distance defined in Eq.~\ref{eq:final_distance} for computing TETOT. TETOT with small label cost achieves the best correlation with transferability.
      \label{table:effect_of_lambda}
    }
    \resizebox{0.34\textwidth}{!}{
    \begin{tabular}{|c|cccc|}
      \hline
     Dataset & 0 & 1 & 1E2 & 1E4 \\
     \hline
     PACS & -0.74 & {\bf -0.76} & -0.74 & -0.65 \\
     VLCS & -0.24 & {\bf-0.25} & {\bf-0.25} & -0.24 \\
     \hline
     Average & -0.49 & {\bf -0.51} & -0.50 & -0.44 \\
      \hline

    \end{tabular}
    }
  \end{center}

\end{table}

\section{Conclusion}
In this work, we proposed TETOT, an efficiently computable metric to gauge a model's performance at test time. 
Our metric can estimate transferability only using the information available at test-time, which includes the knowledge of the source data (or its statistics), parameters of various pre-trained models, and unlabeled data from the target domain. 
Using these, TETOT computes the distributional divergence between the distributions of the source and target domain using Optimal Transport. 
We showed the effectiveness of TETOT on various practical applications such as architecture selection, source selection, and predicting the performance of unseen domains.
We used PACS and VLCS along with their corrupted versions to demonstrate that TETOT achieves a high (negative) correlation with transferability and significantly outperforms the competitive prediction entropy-based metric in all applications.
Our results demonstrated the utility of TETOT in estimating the performance of models, at test time, on data from unseen domains.

\section{Acknowledgment}
This work was partly supported by the NSF EPSCoR-Louisiana Materials Design Alliance (LAMDA) program \#OIA-1946231 and partly by the Harold L. and Heather E. Jurist Center of Excellence for Artificial Intelligence at Tulane University.

{
    \small
    \bibliographystyle{ieeenat_fullname}
    \bibliography{main}

\begin{thebibliography}{42}
\providecommand{\natexlab}[1]{#1}
\providecommand{\url}[1]{\texttt{#1}}
\expandafter\ifx\csname urlstyle\endcsname\relax
  \providecommand{\doi}[1]{doi: #1}\else
  \providecommand{\doi}{doi: \begingroup \urlstyle{rm}\Url}\fi

\bibitem[Albuquerque et~al.(2019)Albuquerque, Monteiro, Darvishi, Falk, and Mitliagkas]{albuquerque2019generalizing}
Isabela Albuquerque, Jo{\~a}o Monteiro, Mohammad Darvishi, Tiago~H Falk, and Ioannis Mitliagkas.
\newblock Generalizing to unseen domains via distribution matching.
\newblock \emph{arXiv preprint arXiv:1911.00804}, 2019.

\bibitem[Alvarez-Melis and Fusi(2020)]{alvarez2020geometric}
David Alvarez-Melis and Nicolo Fusi.
\newblock Geometric dataset distances via optimal transport.
\newblock \emph{arXiv preprint arXiv:2002.02923}, 2020.

\bibitem[Bao et~al.(2019)Bao, Li, Huang, Zhang, Zheng, Zamir, and Guibas]{8803726}
Yajie Bao, Yang Li, Shao-Lun Huang, Lin Zhang, Lizhong Zheng, Amir Zamir, and Leonidas Guibas.
\newblock An information-theoretic approach to transferability in task transfer learning.
\newblock In \emph{2019 IEEE International Conference on Image Processing (ICIP)}, pages 2309--2313, 2019.

\bibitem[Ben-David et~al.(2007)Ben-David, Blitzer, Crammer, Pereira, et~al.]{ben2007analysis}
Shai Ben-David, John Blitzer, Koby Crammer, Fernando Pereira, et~al.
\newblock Analysis of representations for domain adaptation.
\newblock \emph{Advances in neural information processing systems}, 19:\penalty0 137, 2007.

\bibitem[Ben-David et~al.(2010)Ben-David, Blitzer, Crammer, Kulesza, Pereira, and Vaughan]{ben2010theory}
Shai Ben-David, John Blitzer, Koby Crammer, Alex Kulesza, Fernando Pereira, and Jennifer~Wortman Vaughan.
\newblock A theory of learning from different domains.
\newblock \emph{Machine learning}, 79\penalty0 (1):\penalty0 151--175, 2010.

\bibitem[Bulusu et~al.(2020)Bulusu, Kailkhura, Li, Varshney, and Song]{bulusu2020anomalous}
Saikiran Bulusu, Bhavya Kailkhura, Bo Li, Pramod~K Varshney, and Dawn Song.
\newblock Anomalous example detection in deep learning: A survey.
\newblock \emph{IEEE Access}, 8:\penalty0 132330--132347, 2020.

\bibitem[Courty et~al.(2017)Courty, Flamary, Habrard, and Rakotomamonjy]{courty2017joint}
Nicolas Courty, R{\'e}mi Flamary, Amaury Habrard, and Alain Rakotomamonjy.
\newblock Joint distribution optimal transportation for domain adaptation.
\newblock \emph{Advances in Neural Information Processing Systems}, 30, 2017.

\bibitem[Cuturi(2013)]{cuturi2013sinkhorn}
Marco Cuturi.
\newblock Sinkhorn distances: Lightspeed computation of optimal transport.
\newblock \emph{Advances in neural information processing systems}, 26, 2013.

\bibitem[Damodaran et~al.(2018)Damodaran, Kellenberger, Flamary, Tuia, and Courty]{damodaran2018deepjdot}
Bharath~Bhushan Damodaran, Benjamin Kellenberger, R{\'e}mi Flamary, Devis Tuia, and Nicolas Courty.
\newblock Deepjdot: Deep joint distribution optimal transport for unsupervised domain adaptation.
\newblock In \emph{Proceedings of the European Conference on Computer Vision (ECCV)}, pages 447--463, 2018.

\bibitem[Fang et~al.(2013)Fang, Xu, and Rockmore]{fang2013unbiased}
Chen Fang, Ye Xu, and Daniel~N Rockmore.
\newblock Unbiased metric learning: On the utilization of multiple datasets and web images for softening bias.
\newblock In \emph{Proceedings of the IEEE International Conference on Computer Vision}, pages 1657--1664, 2013.

\bibitem[Flamary et~al.(2021)Flamary, Courty, Gramfort, Alaya, Boisbunon, Chambon, Chapel, Corenflos, Fatras, Fournier, Gautheron, Gayraud, Janati, Rakotomamonjy, Redko, Rolet, Schutz, Seguy, Sutherland, Tavenard, Tong, and Vayer]{flamary2021pot}
R{\'e}mi Flamary, Nicolas Courty, Alexandre Gramfort, Mokhtar~Z. Alaya, Aur{\'e}lie Boisbunon, Stanislas Chambon, Laetitia Chapel, Adrien Corenflos, Kilian Fatras, Nemo Fournier, L{\'e}o Gautheron, Nathalie~T.H. Gayraud, Hicham Janati, Alain Rakotomamonjy, Ievgen Redko, Antoine Rolet, Antony Schutz, Vivien Seguy, Danica~J. Sutherland, Romain Tavenard, Alexander Tong, and Titouan Vayer.
\newblock Pot: Python optimal transport.
\newblock \emph{Journal of Machine Learning Research}, 22\penalty0 (78):\penalty0 1--8, 2021.

\bibitem[Ganin et~al.(2016)Ganin, Ustinova, Ajakan, Germain, Larochelle, Laviolette, Marchand, and Lempitsky]{ganin2016domain}
Yaroslav Ganin, Evgeniya Ustinova, Hana Ajakan, Pascal Germain, Hugo Larochelle, Fran{\c{c}}ois Laviolette, Mario Marchand, and Victor Lempitsky.
\newblock Domain-adversarial training of neural networks.
\newblock \emph{The journal of machine learning research}, 17\penalty0 (1):\penalty0 2096--2030, 2016.

\bibitem[Gulrajani and Lopez-Paz(2020)]{gulrajani2020search}
Ishaan Gulrajani and David Lopez-Paz.
\newblock In search of lost domain generalization.
\newblock \emph{arXiv preprint arXiv:2007.01434}, 2020.

\bibitem[Hendrycks and Dietterich(2019)]{hendrycks2019robustness}
Dan Hendrycks and Thomas Dietterich.
\newblock Benchmarking neural network robustness to common corruptions and perturbations.
\newblock \emph{Proceedings of the International Conference on Learning Representations}, 2019.

\bibitem[Hendrycks and Dietterich(2018)]{hendrycks2018benchmarking}
Dan Hendrycks and Thomas~G Dietterich.
\newblock Benchmarking neural network robustness to common corruptions and surface variations.
\newblock \emph{arXiv preprint arXiv:1807.01697}, 2018.

\bibitem[Hoffman et~al.(2018)Hoffman, Tzeng, Park, Zhu, Isola, Saenko, Efros, and Darrell]{hoffman2018cycada}
Judy Hoffman, Eric Tzeng, Taesung Park, Jun-Yan Zhu, Phillip Isola, Kate Saenko, Alexei Efros, and Trevor Darrell.
\newblock Cycada: Cycle-consistent adversarial domain adaptation.
\newblock In \emph{International conference on machine learning}, pages 1989--1998. PMLR, 2018.

\bibitem[Huang et~al.(2022)Huang, Wei, Rong, Yang, and Huang]{huang2022frustratingly}
Long-Kai Huang, Ying Wei, Yu Rong, Qiang Yang, and Junzhou Huang.
\newblock Frustratingly easy transferability estimation, 2022.

\bibitem[Iwasawa and Matsuo(2021)]{iwasawa2021test}
Yusuke Iwasawa and Yutaka Matsuo.
\newblock Test-time classifier adjustment module for model-agnostic domain generalization.
\newblock \emph{Advances in Neural Information Processing Systems}, 34:\penalty0 2427--2440, 2021.

\bibitem[Johansson et~al.(2019)Johansson, Sontag, and Ranganath]{johansson2019support}
Fredrik~D Johansson, David Sontag, and Rajesh Ranganath.
\newblock Support and invertibility in domain-invariant representations.
\newblock In \emph{The 22nd International Conference on Artificial Intelligence and Statistics}, pages 527--536. PMLR, 2019.

\bibitem[Li et~al.(2017)Li, Yang, Song, and Hospedales]{li2017deeper}
Da Li, Yongxin Yang, Yi-Zhe Song, and Timothy~M Hospedales.
\newblock Deeper, broader and artier domain generalization.
\newblock In \emph{Proceedings of the IEEE international conference on computer vision}, pages 5542--5550, 2017.

\bibitem[Liang et~al.(2020)Liang, Hu, and Feng]{liang2020we}
Jian Liang, Dapeng Hu, and Jiashi Feng.
\newblock Do we really need to access the source data? source hypothesis transfer for unsupervised domain adaptation.
\newblock In \emph{International conference on machine learning}, pages 6028--6039. PMLR, 2020.

\bibitem[Long et~al.(2018)Long, Cao, Wang, and Jordan]{long2018conditional}
Mingsheng Long, Zhangjie Cao, Jianmin Wang, and Michael~I Jordan.
\newblock Conditional adversarial domain adaptation.
\newblock \emph{Advances in neural information processing systems}, 31, 2018.

\bibitem[Mansour et~al.(2009)Mansour, Mohri, and Rostamizadeh]{mansour2009domain}
Yishay Mansour, Mehryar Mohri, and Afshin Rostamizadeh.
\newblock Domain adaptation: Learning bounds and algorithms.
\newblock \emph{arXiv preprint arXiv:0902.3430}, 2009.

\bibitem[Mehra et~al.(2021)Mehra, Kailkhura, Chen, and Hamm]{mehra2021understanding}
Akshay Mehra, Bhavya Kailkhura, Pin-Yu Chen, and Jihun Hamm.
\newblock Understanding the limits of unsupervised domain adaptation via data poisoning.
\newblock In \emph{Thirty-Fifth Conference on Neural Information Processing Systems}, 2021.

\bibitem[Mehra et~al.(2022)Mehra, Kailkhura, Chen, and Hamm]{mehra2022do}
Akshay Mehra, Bhavya Kailkhura, Pin-Yu Chen, and Jihun Hamm.
\newblock Do domain generalization methods generalize well?
\newblock In \emph{NeurIPS ML Safety Workshop}, 2022.

\bibitem[Mehra et~al.(2023)Mehra, Zhang, and Hamm]{mehra2023analysis}
Akshay Mehra, Yunbei Zhang, and Jihun Hamm.
\newblock Analysis of task transferability in large pre-trained classifiers.
\newblock \emph{arXiv preprint arXiv:2307.00823}, 2023.

\bibitem[Mehra et~al.(2024)Mehra, Zhang, Kailkhura, and Hamm]{mehra2024fly}
Akshay Mehra, Yunbei Zhang, Bhavya Kailkhura, and Jihun Hamm.
\newblock On the fly neural style smoothing for risk-averse domain generalization.
\newblock In \emph{Proceedings of the IEEE/CVF Winter Conference on Applications of Computer Vision}, pages 3800--3811, 2024.

\bibitem[Mummadi et~al.(2021)Mummadi, Hutmacher, Rambach, Levinkov, Brox, and Metzen]{mummadi2021test}
Chaithanya~Kumar Mummadi, Robin Hutmacher, Kilian Rambach, Evgeny Levinkov, Thomas Brox, and Jan~Hendrik Metzen.
\newblock Test-time adaptation to distribution shift by confidence maximization and input transformation.
\newblock \emph{arXiv preprint arXiv:2106.14999}, 2021.

\bibitem[Nguyen et~al.(2020)Nguyen, Hassner, Seeger, and Archambeau]{nguyen2020leep}
Cuong~V. Nguyen, Tal Hassner, Matthias Seeger, and Cedric Archambeau.
\newblock Leep: A new measure to evaluate transferability of learned representations, 2020.

\bibitem[Qiao et~al.(2020)Qiao, Zhao, and Peng]{qiao2020learning}
Fengchun Qiao, Long Zhao, and Xi Peng.
\newblock Learning to learn single domain generalization.
\newblock In \emph{Proceedings of the IEEE/CVF Conference on Computer Vision and Pattern Recognition}, pages 12556--12565, 2020.

\bibitem[Shen et~al.(2018)Shen, Qu, Zhang, and Yu]{shen2018wasserstein}
Jian Shen, Yanru Qu, Weinan Zhang, and Yong Yu.
\newblock Wasserstein distance guided representation learning for domain adaptation.
\newblock In \emph{Thirty-Second AAAI Conference on Artificial Intelligence}, 2018.

\bibitem[Tan et~al.(2021)Tan, Li, and Huang]{tan2021otce}
Yang Tan, Yang Li, and Shao-Lun Huang.
\newblock Otce: A transferability metric for cross-domain cross-task representations, 2021.

\bibitem[Thopalli et~al.(2023)Thopalli, Turaga, and Thiagarajan]{thopalli2023domain}
Kowshik Thopalli, Pavan Turaga, and Jayaraman~J Thiagarajan.
\newblock Domain alignment meets fully test-time adaptation.
\newblock In \emph{Asian Conference on Machine Learning}, pages 1006--1021. PMLR, 2023.

\bibitem[Tran et~al.(2019)Tran, Nguyen, and Hassner]{tran2019transferability}
Anh~T. Tran, Cuong~V. Nguyen, and Tal Hassner.
\newblock Transferability and hardness of supervised classification tasks, 2019.

\bibitem[Villani(2009)]{villani2009optimal}
C{\'e}dric Villani.
\newblock \emph{Optimal transport: old and new}.
\newblock Springer, 2009.

\bibitem[Volpi et~al.(2018)Volpi, Namkoong, Sener, Duchi, Murino, and Savarese]{volpi2018generalizing}
Riccardo Volpi, Hongseok Namkoong, Ozan Sener, John~C Duchi, Vittorio Murino, and Silvio Savarese.
\newblock Generalizing to unseen domains via adversarial data augmentation.
\newblock \emph{Advances in neural information processing systems}, 31, 2018.

\bibitem[Wang et~al.(2020)Wang, Shelhamer, Liu, Olshausen, and Darrell]{wang2020tent}
Dequan Wang, Evan Shelhamer, Shaoteng Liu, Bruno Olshausen, and Trevor Darrell.
\newblock Tent: Fully test-time adaptation by entropy minimization.
\newblock \emph{arXiv preprint arXiv:2006.10726}, 2020.

\bibitem[Wang et~al.(2021)Wang, Lan, Liu, Ouyang, Zeng, and Qin]{wang2021generalizing}
Jindong Wang, Cuiling Lan, Chang Liu, Yidong Ouyang, Wenjun Zeng, and Tao Qin.
\newblock Generalizing to unseen domains: A survey on domain generalization.
\newblock \emph{arXiv preprint arXiv:2103.03097}, 2021.

\bibitem[You et~al.(2021)You, Liu, Wang, and Long]{you2021logme}
Kaichao You, Yong Liu, Jianmin Wang, and Mingsheng Long.
\newblock Logme: Practical assessment of pre-trained models for transfer learning, 2021.

\bibitem[Zhang et~al.(2021)Zhang, Zhao, Yu, and Poupart]{zhang2021quantifying}
Guojun Zhang, Han Zhao, Yaoliang Yu, and Pascal Poupart.
\newblock Quantifying and improving transferability in domain generalization.
\newblock \emph{arXiv preprint arXiv:2106.03632}, 2021.

\bibitem[Zhao et~al.(2018)Zhao, Zhang, Wu, Moura, Costeira, and Gordon]{zhao2018adversarial}
Han Zhao, Shanghang Zhang, Guanhang Wu, Jos{\'e}~MF Moura, Joao~P Costeira, and Geoffrey~J Gordon.
\newblock Adversarial multiple source domain adaptation.
\newblock \emph{Advances in neural information processing systems}, 31, 2018.

\bibitem[Zhao et~al.(2019)Zhao, Des~Combes, Zhang, and Gordon]{zhao2019learning}
Han Zhao, Remi~Tachet Des~Combes, Kun Zhang, and Geoffrey Gordon.
\newblock On learning invariant representations for domain adaptation.
\newblock In \emph{International Conference on Machine Learning}, pages 7523--7532. PMLR, 2019.

\end{thebibliography}
}


\end{document}